\documentclass[a4paper,preprint,fleqn]{cas-sc}

\usepackage[authoryear,longnamesfirst]{natbib}

\def\tsc#1{\csdef{#1}{\textsc{\lowercase{#1}}\xspace}}
\tsc{WGM}
\tsc{QE}
\usepackage[utf8]{inputenc}
\usepackage[T1]{fontenc}
\usepackage{amsfonts}
\usepackage{microtype}
\usepackage{xcolor}
\usepackage{graphicx}
\usepackage{amsmath}
\usepackage{amssymb}
\usepackage{mathtools}
\usepackage{amsthm}
\usepackage{amsmath}
\usepackage{float}
\usepackage{extarrows}
\usepackage{stmaryrd}
\usepackage{soul}
\usepackage{mathrsfs}

\usepackage{multirow}
\usepackage{array}
\usepackage{cancel}
\usepackage{physics}
\usepackage{subcaption}
\usepackage{xspace}
\usepackage{caption}
\usepackage{adjustbox}
\usepackage{amssymb} 
\usepackage{amsfonts}

\usepackage{float}

\usepackage{tikz}
\usetikzlibrary{positioning,shapes.geometric,arrows.meta}

\usepackage{mdframed}
\usepackage{pgf} % or tikz

\usepackage{enumitem}

\usepackage{array}     % for advanced column formatting
\usepackage{arydshln}  % for dashed lines (\hdashline, \cdashline)

\theoremstyle{plain}
\newtheorem{theorem}{Theorem}[section]

\newtheorem{lemma}[theorem]{Lemma}
\newtheorem{corollary}[theorem]{Corollary}
\theoremstyle{definition}
\newtheorem{definition}[theorem]{Definition}
\newtheorem{assumption}[theorem]{Assumption}

\theoremstyle{remark}
\newtheorem{remark}[theorem]{Remark}
\newtheorem{mynewproof}[theorem]{Proof}

\newcommand{\ie}{\emph{i.e.}}
\newcommand{\eg}{\emph{e.g.}}

\graphicspath{{./}{./plots/}}

\makeatletter
\def\input@path{{./}{./tikz/}}
\makeatother

\begin{document}

\let\WriteBookmarks\relax
\def\floatpagepagefraction{1}
\def\textpagefraction{.001}

\shorttitle{Higher-Order Singular-Value Derivatives of Real Rectangular Matrices}    

\shortauthors{R\'ois\'in Luo et al.}

\title [mode = title]{Higher-Order Singular-Value Derivatives of Real Rectangular Matrices}




\author[1,2]{R\'ois\'in Luo}[orcid=0000-0002-5365-0379]

%\fntext[1]{R\'ois\'in is a Ph.D candidate in artificial intellgience.}

\ead{roisincrtai@gmail.com}



\affiliation[1]{organization={University of Galway},
            %addressline={University Road}, 
            %city={Galway},
%          citysep={}, % Uncomment if no comma needed between city and postcode
            %postcode={H91 TK33}, 
            %state={},
            country={Ireland}}

\affiliation[2]{organization={Research Ireland -- Centre for Research Training in Artificial Intelligence (CRT-AI)},
            %addressline={University Road}, 
            %city={Galway},
%          citysep={}, % Uncomment if no comma needed between city and postcode
            %postcode={H91 TK33}, 
            %state={},
            country={Ireland}
            }

\author[1,2]{Colm O'Riordan}%[]

\author[1,2]{James McDermott}%[]

\cormark[1]

\cortext[1]{Corresponding author}

% Footnote of the second author

% Email id of the second author
%\ead{}



% \affiliation[2]{organization={},
% %          citysep={}, % Uncomment if no comma needed between city and postcode

\fntext[1]{Reproducibility for numerical experiments: \url{https://github.com/roisincrtai/highorder_spectral_variation_analysis}}


\begin{abstract}
Higher-order derivatives of singular values in real rectangular matrices arise naturally in both numerical simulation and theoretical analysis, with applications in areas such as statistical physics and optimization in deep learning. Deriving closed-form expressions beyond first order has remained a difficult problem within classical matrix analysis, and no general framework has been available. To address this gap, we present an operator-theoretic framework that extends Kato's analytic perturbation theory from self-adjoint operators to real rectangular matrices, thereby yielding general $n$-th order Fr\'echet derivatives of singular values. As a special case, we obtain a closed-form Kronecker-product representation of the singular-value Hessian, not previously found in the literature. This framework bridges abstract perturbation theory with matrix analysis and provides a systematic tool for higher-order spectral analysis. %\textbf{Code: \url{https://github.com/roisincrtai/highorder_spectral_variation_analysis}}

%old version
%We present a theoretical framework for deriving the general $n$-th order Fr\'echet derivatives of singular values in real rectangular matrices, by leveraging reduced resolvent operators from Kato's analytic perturbation theory for self‐adjoint operators. Deriving closed‐form expressions for higher‐order derivatives of singular values is notoriously challenging through standard matrix‐analysis techniques. To overcome this, we treat a real rectangular matrix as a compact operator on a finite‐dimensional Hilbert space, and embed the rectangular matrix into a block self‐adjoint operator so that non‐symmetric perturbations are captured. Applying Kato's asymptotic eigenvalue expansion to this construction, we obtain a general, closed‐form expression for the infinitesimal $n$-th order spectral variations. Specializing to $n=2$ and deploying on a Kronecker‐product representation with matrix convention yield the Hessian of a singular value, not found in literature. By bridging abstract operator‐theoretic perturbation theory with matrices, our framework equips researchers with a practical toolkit for higher‐order spectral sensitivity studies in random matrix applications (\eg~adversarial perturbation in deep learning). \textbf{Code: \url{https://github.com/roisincrtai/highorder_spectral_variation_analysis}}

\end{abstract}

% \includegraphics{}

% \begin{highlights}
% \item 
% \item 
% \item 
% \end{highlights}

% % Keywords
% % Each keyword is seperated by \sep

\maketitle

\section{Introduction}

Singular values lie at the core of modern matrix analysis, encapsulating key spectral information such as \emph{operator norm}, \emph{conditioning}, \emph{effective rank}, and underpinning applications across numerical linear algebra, data science, control theory, and mathematical physics \citep{edelman2005random,yosida2012functional,Horn2012,tao2012topics}. In random matrix theory, singular values govern limiting laws such as Marchenko--Pastur distributions \citep{MarcenkoPastur1967}, edge fluctuations described by Tracy--Widom laws \citep{TracyWidom1994}, and fine-scale local statistics such as local eigenvalue spacings \citep{Mehta2004}. In physics and deep learning, higher-order derivatives of singular values are indispensable for rigorous analysis in stochastic dynamical settings, where systems are subject to noise and random perturbations \citep{oksendal2003stochastic}. 

For example, let $\boldsymbol{\theta}_t \in \mathbb{R}^{m \times n}$ be a parameter matrix with $r$ non-zero singular values $\sigma_1,\sigma_2,\ldots,\sigma_r$. Its dynamics are characterized by an adapted It\^o process \citep{ito1951stochastic}
\begin{align}
    \dd \mathrm{vec}(\boldsymbol{\theta}_t) 
    = G_t \dd t 
      + D_t \dd W_t,
\end{align}
where $\mathrm{vec}(\cdot)$ denotes the vectorization operator, $G_t \in \mathbb{R}^{mn}$ is the \emph{drift term}, $D_t \in \mathbb{R}^{mn \times mn}$ is the \emph{diffusion coefficient}, and $\dd W_t$ is a high-dimensional Wiener process in $\mathbb{R}^{mn}$. Let
\begin{align}
\phi_t = \phi(\sigma_1,\sigma_2,\ldots,\sigma_r)(t)    
\end{align}
denote a spectral functional of the singular values of $\boldsymbol{\theta}_t$, then applying It\^o's lemma shows that the rigorous analysis of the spectral dynamics of $\dd \phi_t$ requires the second-order derivatives of singular values \citep{oksendal2003stochastic}. Such induced dynamics arise naturally in both physics and deep learning. In physics, the \emph{von Neumann entropy} --- a measure of the statistical uncertainty within a quantum system --- is a spectral functional of singular values \citep{fano1957description}, and widely used in the study of quantum entanglement \citep{nielsen2010quantum}. In deep learning, the Lipschitz continuity of neural networks is a spectral functional of the largest singular value \citep{luo2025lipschitz}. For more general non-Gaussian drivers in stochastic dynamics, such as L\'evy processes, higher-order derivatives of singular values are indispensable for rigorous analysis and for deriving sharp bounds \citep{applebaum2009levy}.

%challenges
Although first-order derivatives of singular values are well known in the literature \citep{Wedin1972,StewartSun1990,strang2012linear}, explicit closed-form expressions for second- and higher-order derivatives are largely absent from the literature. A unified, highly procedural, and systematic framework for their derivation has been lacking, since direct approaches via matrix analysis are challenging due to the intricate interplay among local spectral structures (\eg, spectral gaps), left and right singular subspaces, and the associated null spaces.

To bridge this gap, we present an operator-theoretic framework for deriving arbitrary higher-order derivatives of singular values in a highly procedural approach. Our approach treats matrices as bounded linear operators on Hilbert spaces and extends Kato's analytic perturbation theory \citep{kato1995perturbation} beyond the self-adjoint setting. The key step is to embed a non-self-adjoint real rectangular matrix into a self-adjoint operator via the Jordan–Wielandt embedding (\ie, Hermitian dilation trick) \citep{wielandt1955eigenvalues,StewartSun1990,shalit2021dilation}. We then analyze the asymptotic expansions of the resulting eigenvalues by extending Kato's results in eigenvalue expansions, relate these eigenvalue expansions to Fr\'echet derivatives of singular values, and express the Fr\'echet derivative tensors with Kronecker-product representation.

\subsection{Perturbation Theory}

Classical perturbation theory has developed along several independent traditions. For example, \textbf{analytic operator-theoretic perturbation theory} \citep{rellich1969perturbation,kato1995perturbation} treats holomorphic families of operators on Banach or Hilbert spaces, using resolvents, Riesz projectors, and contour integrals to prove the existence of analytic eigenvalue and eigenspace branches and to derive expansion formulas, including trace identities for eigenvalue clusters. \textbf{Matrix perturbation theory} \citep{StewartSun1990,Horn2012,bhatia2013matrix} focuses on the finite-dimensional case and derives explicit perturbation formulas via algebraic tools such as characteristic polynomials, Schur forms, and Sylvester equations, typically without explicitly invoking the operator-theoretic machinery. \textbf{Rayleigh--Schr\"odinger perturbation theory} \citep{rayleigh1896theory,schrodinger1926quantisierung,sakurai2020modern} in quantum mechanics provides basis-dependent expansions in terms of matrix elements and energy gaps; these coincide with the analytic expansions under discreteness and gap assumptions, but are often presented in physics as formal series rather than within Kato’s framework. Despite their differences in tool and emphasis, these frameworks are mathematically consistent and recover the same perturbative corrections in overlapping regimes.

\begin{figure}%[!t]
    \centering
    
    \resizebox{0.75\linewidth}{!}{
      \begin{tikzpicture}[node distance=10mm and 15mm, >=Stealth,
  kato/.style={  rectangle, 
                draw, 
                rounded corners, 
                fill=cyan!30, 
                align=center, 
                text width=6cm, 
                inner sep=0.2cm, 
                minimum height=3cm, 
          },
  complex/.style={  rectangle, 
                draw, 
                rounded corners, 
                fill=green!30, 
                align=center, 
                text width=6cm, 
                inner sep=0.2cm, 
                minimum height=3cm, 
          }, 
  ours/.style={
                rectangle, 
                draw, 
                rounded corners, 
                fill=orange!30, 
                align=center, 
                text width=6cm,
                inner sep=0.2cm, 
                minimum height=3cm, 
            },
  app/.style={
                rectangle, 
                draw, 
                rounded corners, 
                fill=purple!15, 
                align=center, 
                text width=6cm,
                inner sep=0.2cm, 
                minimum height=3cm, 
            }, 
  legendbox/.style={
    rectangle,
    draw,
    rounded corners,
    minimum width=8mm,
    minimum height=6mm,
    inner sep=0pt,
  },
  legend label/.style={
    anchor=west,
    font=\footnotesize,
  }
]

  %add legend
  \node[legendbox, fill=cyan!30] (legC) {};
  \node[legend label, right=1mm of legC] (labC) {Kato's framework \citep{kato1995perturbation}};

  \node[legendbox, fill=orange!30, right=5cm of legC] (legO) {};
  \node[legend label, right=1mm of legO] (labO) {Our framework};

  \node[legendbox, fill=green!50, right=3cm of legO] (legG) {};
  \node[legend label, right=1mm of legG] (labG) {Functional and complex analysis};

 \begin{scope}[yshift=-20mm]
  \node[kato] (PerturbationSeries) {
    Perturbation Series 
    $$\mathcal{T}(x)=...$$
    (see Section~\ref{sec:eigenvalue_expansion})
  };

  \node[kato, right=of PerturbationSeries] (Resolvent) {
    Resolvent Operator
    $$ R(z) = (\mathcal{T}(x) - zI)^{-1} $$
    (see Section~\ref{sec:eigenvalue_expansion})
  };

  \node[kato, right=of Resolvent] (RieszProjection) {
    Riesz Projector
    $$
    P(x) = -\frac{1}{2\pi\,i}\oint_{\Gamma} R(z) dz
    $$
    where $\Gamma$ is a small contour enclosing only an isolated eigenvalue $\lambda^{(0)}$. \\
    (see Section~\ref{sec:eigenvalue_expansion})
  };
  
  \node[kato, below=of RieszProjection, xshift=0cm] (PerturbationProjection) {
    Perturbation Projection
    \begin{align}
    &\mathcal{T}(x)P(x) = \lambda(x)P(x) \notag \\
    &\Longrightarrow \lambda(x) = ...  \notag\\
    &\Longrightarrow \text{equating coefficient of}~x^n~...\notag
    \end{align}
    (see Section~\ref{sec:eigenvalue_expansion})
  };

  \node[ours, below=of PerturbationProjection] (EigenvalueExpansion) {
    Theorem~\ref{thm:expansions_of_eigenvalues} (Refined Asymptotic Eigenvalue Expansion): Expanding $\lambda(x)$ and applying residue theorem to further simplify yield
    $$\lambda^{(n)}= \cdots$$
    (see Section~\ref{sec:eigenvalue_expansion})
  };

  \node[complex, left=of EigenvalueExpansion, ] (NeumannSeries) {
    Operator Neumann series of $R(z)$\\
    Residue Theorem\\
    $\cdots$\\
    (see Section~\ref{sec:eigenvalue_expansion})
  };

  \node[ours, minimum height=5cm, below=of EigenvalueExpansion] (ApplyEigenvalueExpansion) {
  Theorem~\ref{thm:nth_order_singular_value} (Infinitesimal Spectral Variation): 
    Applying Theorem~\ref{theorem:Jordan_Wielandt_relation} (Jordan--Wielandt Embedding), Theorem~\ref{thm:expansions_of_eigenvalues} (Asymptotic Eigenvalue Expansion), and Theorem~(\ref{thm:frechet_taylor_operator}) (Analytic Perturbation for Holomorphic Operators) on $\mathcal{T}(x)$ produces the higher-order eigenvalue expansion
    $$
    \sigma_k^{(n)}=\cdots %, \quad D^n \sigma_k = n! \sigma_k^{(n)}
    $$
    (see Section~\ref{sec:spectral_variation})
  };

\node[ours, minimum height=4cm, left=of ApplyEigenvalueExpansion] (ConstructTx) {
    % \node[ours, minimum height=4cm, below=of DefineAx] (ConstructTx) {
    Theorem~\ref{theorem:Jordan_Wielandt_relation} (Jordan--Wielandt Embedding): Embed $A$ into Jordan–Wielandt Embedding
    $$
    \mathcal{T}=\begin{bmatrix}
       O & A \\
       A^\top & O
    \end{bmatrix}
    $$
    for capturing perturbations from all subspaces in a rectangular matrix $A$ (see Section~\ref{sec:schematic_overview}).
  };

    \node[ours, minimum height=3cm, left=of ConstructTx] (DefineAx) {
  % \node[ours, minimum height=3cm, below=of PerturbationSeries] (DefineAx) {
    A rectangular matrix $A$ is a non-self-adjoint linear operator.\\
    (see Section~\ref{sec:schematic_overview}).
  };

  \node[ours, below=of ApplyEigenvalueExpansion] (MapToFrechetDerivative) {
    Map the perturbed eigenvalue expansion to Fr\'echet derivative by
    $$
    D^n\sigma_k[\dd A, \cdots] = n!\lim_{x \to 0}x^n\sigma_k^{(n)}
    $$
    (see Section~\ref{sec:spectral_variation})
  };

  \node[ours, left=of MapToFrechetDerivative] (MapToMatrixLayout) {
    Specialize $n$ (\eg~$n=1,2,\cdots$) and map the Fr\'echet derivative
    $$
    D^n\sigma_k[\dd A, \cdots]
    $$
    to matrix layout convention.\\
    (see Section~\ref{sec:special_case_jacobian} and Section~\ref{sec:special_case_hessian})
  };

   \node[ours, left=of MapToMatrixLayout] (LayoutConvention) {
    Fr\'echet Derivative and Layout Convention\\
    (see Section~\ref{sec:layout_convention}).
  };

   \draw[->,thick] (PerturbationSeries) -- (Resolvent);
   \draw[->,thick] (Resolvent)  -- (RieszProjection);
   \draw[->,thick] (Resolvent)  -- (NeumannSeries);
   %\draw[->,thick] (Resolvent)  |- (ReducedResolvent);
   \draw[->,thick] (RieszProjection)  -- (PerturbationProjection);
   \draw[->,thick] (NeumannSeries)  -- (EigenvalueExpansion);
   %\draw[->] (ResidueTheorem)  -- (EigenvalueExpansion);
   \draw[->,thick] (PerturbationProjection) -- (EigenvalueExpansion);
   \draw[->,thick,dotted] (PerturbationSeries) -- (DefineAx);
   \draw[->,thick] (DefineAx) -- (ConstructTx);
   % \draw[->,thick] (ConstructTx) |- (ApplyEigenvalueExpansion);
   \draw[->,thick] (ConstructTx) -- (ApplyEigenvalueExpansion);
   \draw[->,thick] (EigenvalueExpansion) -- (ApplyEigenvalueExpansion);
   \draw[->,thick] (ApplyEigenvalueExpansion) -- (MapToFrechetDerivative);
   %\draw[->,thick] (SimplifyEigenvalueExpansion) -- (MapToFrechetDerivative);
   \draw[->,thick] (MapToFrechetDerivative) -- (MapToMatrixLayout);
   \draw[->,thick] (LayoutConvention) -- (MapToMatrixLayout);
 \end{scope}

\end{tikzpicture}
    }

    \vspace{2.5ex}

    \caption[Theoretical Framework for Infinitesimal Spectral Variations]{\textbf{Theoretical Framework for Infinitesimal Spectral Variations}. We extend Kato's analytic perturbation theory for self-adjoint operators to derive arbitrary-order singular-value derivatives \citep{kato1995perturbation}. For a rectangular matrix $A$, we introduce its Jordan--Wielandt embedding $\mathcal{T}$ (Theorem~\ref{theorem:Jordan_Wielandt_relation}), a block self-adjoint operator that encodes perturbations across all subspaces (\ie, left-singular, right-singular, left-null, and right-null). By extending Kato’s asymptotic eigenvalue expansions to this embedding and expressing them in explicit closed form --- computing and simplifying with residue theorem --- yields the $n$th-order expansions of singular values of $A$. These expansions are then related to Fr\'echet derivatives, given by analytic perturbation theorem (Theorem~\ref{thm:frechet_taylor_operator}). Finally, by specializing to explicit matrix-layout conventions, we obtain a systematic and constructive procedure for computing arbitrary-order singular-value derivatives of rectangular matrices. Our method is highly procedure for deriving arbitrary-order singular-value derivatives.}
    
    \label{fig:framework_of_spectral_analysis}

\end{figure}

\subsection{Schematic Overview}
\label{sec:schematic_overview}

A schematic overview of the framework is illustrated in Figure~\ref{fig:framework_of_spectral_analysis}. To apply Kato's framework for self-adjoint operators, we first embed a non-self-adjoint $A \in \mathbb{R}^{m \times n}$ (since $A \neq A^\top$) into a self-adjoint operator $\mathcal{T}$ using the Jordan–Wielandt embedding (\ie, Hermitian dilation) \citep{wielandt1955eigenvalues,StewartSun1990,edelman2005random,li2005note,bai2010spectral,Horn2012,shalit2021dilation}, taking:
\begin{align}
\mathcal{T} :=
\begin{bmatrix}
        O & A \\
        A^\top & O
\end{bmatrix}
\in \mathbb{R}^{(m+n)\times(m+n)}.
\end{align}
It is immediate that $\mathcal{T}$ is self-adjoint, since:
\begin{align}
    \mathcal{T}^\top
    =
    \begin{bmatrix}
        O & A \\
        A^\top & O
    \end{bmatrix}^\top
    =
    \begin{bmatrix}
        O & A \\
        A^\top & O
    \end{bmatrix}
    = \mathcal{T}.
\end{align}

This embedding preserves the complete information regarding the spectrum of $A$. The spectrum of $A$ is stated in Theorem~\ref{theorem:full_svd}~(\nameref{theorem:full_svd}), and the spectrum of $\mathcal{T}$ relates to the spectral structure of $A$ as stated in Theorem~\ref{theorem:Jordan_Wielandt_relation}~(\nameref{theorem:Jordan_Wielandt_relation}).

\begin{theorem}[Matrix Singular Value Decomposition (Full Form)]
\label{theorem:full_svd}
Let $A \in \mathbb{R}^{m \times n}$ be a real rectangular matrix. Then $A$ admits a full singular value decomposition (SVD) \citep{Horn2012,StewartSun1990} by:
\begin{align}
    A &= U \Sigma V^\top, \\[4pt]
    U &= \bigl[\, U_r \;\; U_0 \,\bigr] \in \mathbb{R}^{m \times m}, 
    \qquad 
    V = \bigl[\, V_r \;\; V_0 \,\bigr] \in \mathbb{R}^{n \times n}, \\[4pt]
\Sigma &=
\left[
\begin{array}{c:c}
\begin{array}{cccc}
\sigma_1 &        &        &        \\
         & \sigma_2 &      &        \\
         &        & \ddots &        \\
         &        &        & \sigma_r
\end{array}
& 
\begin{array}{c}
\; \vcenter{\hbox{$O_{\,r\times (n-r)}$}} \;
\end{array}
\\ \hdashline
\begin{array}{c}
\; \vcenter{\hbox{$O_{\, (m-r)\times r}$}} \;
\end{array}
&
\begin{array}{c}
\; \vcenter{\hbox{$O_{\, (m-r)\times (n-r)}$}} \;
\end{array}
\end{array}
\right],
\end{align}
where $\sigma_1 \geq \sigma_2 \geq \cdots \geq \sigma_r > 0$ are the non-zero singular values,
$U_r \in \mathbb{R}^{m \times r}$ and $V_r \in \mathbb{R}^{n \times r}$ contain the corresponding left and right singular vectors, and 
$U_0$ and $V_0$ span the left and right null spaces of $A$, respectively.     
\end{theorem}

%- Thm 1.2: Note that the SVD exists for any matrix A. Thus, I suggest changing "[...] If A admits a SVD" with something on the line of "Given the SVD of A";
\begin{theorem}[Spectrum of Jordan–Wielandt Embedding]
\label{theorem:Jordan_Wielandt_relation}
The spectrum of $A \in \mathbb{R}^{m \times n}$ and the spectrum of its Jordan–Wielandt embedding \citep{wielandt1955eigenvalues,StewartSun1990,li2005note,bai2010spectral,Horn2012,shalit2021dilation}:
\begin{align}
\mathcal{T} :=
\begin{bmatrix}
        O & A \\
        A^\top & O
\end{bmatrix}
\in \mathbb{R}^{(m+n)\times(m+n)}
,
\end{align}
are directly related. Given the SVD of $A$ as stated in Theorem~\ref{theorem:full_svd}, the spectrum of $A$ relates to the spectrum of $T$ by:
\begin{align}
    \mathcal{T} \left(\frac{1}{\sqrt{2}}
    \begin{bmatrix} u_i \\ v_i \end{bmatrix}
    \right)= \sigma_i \left(\frac{1}{\sqrt{2}}
    \begin{bmatrix} u_i \\ v_i \end{bmatrix}\right),
\end{align}
and:
\begin{align}
\mathcal{T} 
    \left(\frac{1}{\sqrt{2}}
    \begin{bmatrix} u_i \\ -v_i \end{bmatrix}
    \right)
    = -\sigma_i 
    \left(\frac{1}{\sqrt{2}}
    \begin{bmatrix} u_i \\ -v_i \end{bmatrix}
    \right)
    ,
\end{align}
respectively, where the factor $\tfrac{1}{\sqrt{2}}$ ensures normalization and hence orthonormality of the eigenvectors. Thus, each singular value $\sigma_i$ of $A$ corresponds to a pair of eigenvalues:
\begin{align}
    \lambda_i^{(+)} = \sigma_i, \qquad \lambda_i^{(-)} = -\sigma_i,
\end{align}
with eigenvectors constructed directly from the singular vector pair $(u_i,v_i)$. The null spaces are also preserved in this embedding:
\begin{align}
\ker(\mathcal{T})
= \left\{
\begin{bmatrix} u_j \\ 0 \end{bmatrix} 
: u_j \in \ker(A^\top)\right\}
  \oplus
  \left\{
  \begin{bmatrix} 0 \\ v_k \end{bmatrix} 
  : v_k \in \ker(A)\right\}
  .
\end{align}
\end{theorem}

\begin{remark}
The use of the Jordan–Wielandt embedding to transfer results on Hermitian eigenvalues to singular values of rectangular matrices is well-known in the literature \citep{StewartSun1990,edelman2005random,li2005note,bai2010spectral,Horn2012}. For instance, \citeauthor{StewartSun1990} employ the construction in their analysis of singular-value perturbations, using it to extend Weyl–type inequalities \citep{weyl1912asymptotische,franklin2000matrix} and sensitivity bounds from Hermitian eigenvalues to singular values \citep{StewartSun1990}. \citeauthor{li2005note} also use the embedding to transfer perturbation bounds for Hermitian eigenvalues to singular values of rectangular matrices \citep{li2005note}. Similarly, \citeauthor{Horn2012} present the Hermitian dilation as a standard device in matrix analysis for proving variational characterizations and interlacing properties of singular values \citep{Horn2012}. Unlike these works, which use the Hermitian dilation mainly as a device to transfer known eigenvalue results, our framework exploits it to develop explicit operator-theoretic expansions that yield closed-form higher-order Fr\'echet derivatives of singular values.      
\end{remark}

Next, starting from the eigenvalue expansion of reduced resolvent of operator $\mathcal{T}$ and applying the residue theorem to simplify, we derive the asymptotic eigenvalue expansion of $\mathcal{T}$ up to $n$-th order under holomorphic perturbations (Theorem~\ref{thm:expansions_of_eigenvalues}). By relating the $n$-th order term of this expansion with the corresponding $n$-th order Fr\'echet derivative, we obtain explicit expressions for higher-order derivatives of singular values. Finally, we deploy the $n$-th order Fr\'echet derivative with matrix layout conventions. In particular, the first-order case ($n=1$) recovers the well-known Jacobian of singular values; while the second-order case ($n=2$) yields the singular-value Hessian with Kronecker-product representation, which has not appeared previously in the literature. By bridging the abstract operator-theoretic expansions with matrices, our framework provides a toolkit for arbitrary-order singular-value analysis.

\subsection{Contributions}

This paper makes the following contributions:
\begin{enumerate}

    \item \textbf{Spectral Variations in Rectangular Matrices}. We present an operator-theoretic framework for analyzing $n$-th order spectral variations in real rectangular matrices (see Figure~\ref{fig:framework_of_spectral_analysis}). This framework provides a \textbf{systematic procedure} for deriving higher-order derivatives of singular values in real rectangular matrices.

    \item \textbf{Singular-Value Hessian}. Specializing to $n=2$ yields the second-order derivative (Hessian) of singular values, expressed in a Kronecker-product representation that, to the best of our knowledge, has not appeared previously in the literature. This result is particularly essential for analysis of \textbf{induced spectral stochastic dynamics}, where second-order derivatives arise naturally in It\^o calculus for stochastic differential equations (SDEs) driven by Wiener processes.
    
\end{enumerate}

\section{Fr\'echet Derivative and Layout Convention}
\label{sec:layout_convention}

Deploying results from abstract operator theory in matrix settings requires explicit layout conventions, particularly for the representation of derivatives. Before commencing the theoretical analysis, this section introduces the conventions fundamental to our framework. Section~\ref{sec:matrix_and_svd} introduces matrix layout and the differentiability condition; Section~\ref{sec:matrix_frechet_derivative} presents general Fr\'echet derivatives for matrix-to-matrix maps together with their tensor representations; and Section~\ref{sec:matrix_functional} specializes to Fr\'echet derivatives of matrix-to-scalar functionals and their vectorized Kronecker-product representation \citep{kolda2009tensor}.

\subsection{Matrix and Spectral Decomposition}
\label{sec:matrix_and_svd}

Let
\begin{align}
    A = \begin{bmatrix}
            A_{1,1} & A_{1,2} & \cdots & A_{1,n} \\
            A_{2,1} & A_{2,2} & \cdots & A_{2,n} \\
            \vdots  & \vdots  & \ddots & \vdots  \\
            A_{m,1} & A_{m,2} & \cdots & A_{m,n}
        \end{bmatrix}
        \in \mathbb{R}^{m \times n}
\end{align}
be a real rectangular matrix of rank $r=\mathrm{rank}(A)$, where $A_{i,j}$ denotes its $(i,j)$-th entry. The $A$ admits a \emph{full} SVD as stated in Theorem~\ref{theorem:full_svd}. Specially, the \emph{reduced} or \emph{truncated} SVD of $A$ is given as:
\begin{align}
    A =\sum_{k=1}^{r} \sigma_k u_k v_k^\top,
\end{align}
where $r=\rank(A)$, and $u_k$ and $v_k$ are the left and right singular vectors associated with singular value $\sigma_k > 0$. 

\begin{lemma}[Essential Matrix Identities]
\label{lemma:essential_matrix_identities}
Let $x \in \mathbb{R}$ be a scalar, and real matrices $A$, $B$, $C$ and $V$ be of such sizes that one can form their products. Then the following identities hold \citep{franklin2000matrix,Horn2012,liu2024professor}:
\begin{enumerate}
    \item $\operatorname{vec}(x) = x$,
    \item $\tr(x) = x$,
    \item $\operatorname{vec}(B V A^\top) = (A \otimes B)\operatorname{vec}(V)$,
    \item $(A \otimes B)^\top = A^\top \otimes B^\top$,
    \item $\tr(ABC) = \tr(CAB) = \tr(BCA)$
    .
\end{enumerate}
\end{lemma}

\subsection{Differentiability Condition}

To ensure the existence of higher-order differentiability of non-zero singular values and associated singular vectors, we further assume that the non-zero singular values of $A \in \mathbb{R}^{m \times n}$ are simple (\ie, each non-zero singular value has multiplicity one), as stated in Assumption~\ref{assump:simple_singular_values}~(\nameref{assump:simple_singular_values}). This \emph{simplicity} assumption is essential for ensuring that non-zero singular value $\sigma_i > 0$ of $A$ and associated singular vectors $u_i$ and $v_i$ depend smoothly on the entries of $A$, in fact yielding $u_i, v_i, \sigma_i \in C^{\infty}$ (\ie~maps are infinitely continuously differentiable). Under this assumption, non-zero singular values and their associated singular vectors vary smoothly with perturbations of $A$.

\begin{assumption}[Simplicity Assumption of Non-Zero Singular Values]
\label{assump:simple_singular_values}
We assume that the non-zero singular values of $A$ are \emph{simple}, \ie,
\begin{align}
    \sigma_i\neq\sigma_j \quad \text{for all} \quad i\neq j
    ,
\end{align}
\citep{kato1995perturbation,Horn2012}.
\end{assumption}

If this assumption fails, a non-zero singular value may have multiplicity greater than one; singular values then remain continuous but may fail to be differentiable at points of multiplicity, and the associated singular subspaces are well defined whereas individual singular vectors are not unique. In such settings, higher-order derivatives generally do not exist in the classical context, and analysis must instead be carried out in terms of spectral projectors or within the framework of subdifferential calculus \citep{clarke1990optimization,lewis2005nonsmooth}.

\subsection{Matrix Fr\'echet Derivative as Multilinear Operator}
\label{sec:matrix_frechet_derivative}

We regard matrix Fr\'echet derivatives as multilinear operators \citep{Rudin1991,yosida2012functional}. A definition for general Fr\'echet differentiable real matrix-to-matrix maps and their tensor representation are in Definition~\ref{def:frechet_derivative}~(\nameref{def:frechet_derivative}). The existence and uniqueness of the Fr\'echet derivative are stated in Theorem~\ref{thm:unique-frechet}~(\nameref{thm:unique-frechet}).

\begin{definition}[$\alpha$-Times Continuously Fr\'echet Differentiable Matrix Map]
\label{def:frechet_derivative}
Let
\begin{align}
  F:\,\mathbb{R}^{m\times n}\to\mathbb{R}^{s\times t}
\end{align}
be $\alpha$-times continuously Fr\'echet differentiable (\ie, $F\in C^\alpha$) \citep{Rudin1991,spivak2018calculus,Horn2012}. The $\alpha$-th Fr\'echet derivative of $F$ is a multilinear map:
\begin{align}
D^\alpha F:\;(\mathbb{R}^{m\times n})^\alpha \to \mathbb{R}^{s\times t}.
\end{align}

Writing $F_{i,j}$ for the $(i,j)$-th component of $F$ and $A_{p,q}$ for the $(p,q)$-th entry of $A$, with:
\begin{align}
1\leq i \leq s,\qquad 1\leq j \leq t,\qquad 1\leq p \leq m,\qquad 1\leq q \leq n,
\end{align}
then the $\alpha$-th derivative $D^\alpha F$ at matrix $A$ is a tensor, defined by:
\begin{align}
\bigl[D^\alpha F(A)\bigr]_{i,j\,;\,p_1 q_1\,\dots\,p_\alpha q_\alpha}
 \;=\;
 \frac{\partial^\alpha F_{i,j}(A)}{\partial A_{p_1 q_1}\cdots \partial A_{p_\alpha q_\alpha}}
 \;\in\; \mathbb{R}.
\end{align}

The action of tensor $D^\alpha F(A)$ on directions $H_1,\dots,H_\alpha \in \mathbb{R}^{m\times n}$ is obtained \emph{component-wise} by contracting tensor $D^\alpha F(A)$ with the indices on $H_1,\dots,H_\alpha$:
\begin{align}
\bigl[D^\alpha F(A)[H_1,\dots,H_\alpha]\bigr]_{i,j}
 \;=\;
 \frac{\partial^\alpha F_{i,j}(A)}{\partial A_{p_1 q_1}\cdots \partial A_{p_\alpha q_\alpha}}
 \,
 (H_1)_{p_1 q_1}\cdots (H_\alpha)_{p_\alpha q_\alpha}.
\end{align}

Moreover, for $H\in\mathbb{R}^{m\times n}$, the $F$ at $A$ admits a multivariate Taylor expansion:
\begin{align}
F(A+H)
  = \sum_{\beta=0}^{\alpha} \frac{1}{\beta!}\, D^\beta F(A)[\underbrace{H,\dots,H}_{\beta\ \text{times}}]
    + o\bigl(\|H\|^\alpha\bigr),\qquad (\|H\|\to 0),
\end{align}
where $\|\cdot\|$ is any norm on $\mathbb{R}^{m\times n}$ (\eg, the Frobenius norm).
\end{definition}

\begin{theorem}[Uniqueness of $\alpha$-Times Fr\'echet Derivative \citep{Rudin1991,spivak2018calculus}]
\label{thm:unique-frechet}
Suppose $F \in C^{\alpha}$ is differentiable up to order $\alpha$. Then $D^{\alpha} F$ exists, is a symmetric $\alpha$-linear map, and is \emph{unique}.  That is, there is no other $\alpha $-linear operator satisfying the defining Taylor‐remainder condition. This theorem ensures the uniqueness of the derivatives of singular values under the differentiability condition, as stated in Assumption~\ref{assump:simple_singular_values}~(\nameref{assump:simple_singular_values}).
\end{theorem}

\subsection{Representation Convention for Matrix-Valued Functionals}
\label{sec:matrix_functional}

We focus on the derivatives of singular values, which are matrix-valued functionals. To obtain matrix representations to facilitate concrete applications, we specialize the general matrix-to-matrix maps of Definition~\ref{def:frechet_derivative} to matrix-valued functionals. In general, the $\alpha$-th Fr\'echet derivative is a higher-order tensor. To express such tensors in matrix form, we employ vectorization (with a \textbf{column-major convention}) together with the Kronecker-product representation \citep{kolda2009tensor,magnusmatrix}, as established in Corollary~\ref{corollary:vectorized_kronecker_product_representation}~(\nameref{corollary:vectorized_kronecker_product_representation}). As complementary conventions, we also introduce explicit matrix layouts for the Jacobian in Section~\ref{sec:jacobian_layout}~(\nameref{sec:jacobian_layout}) and for the Hessian in Section~\ref{sec:hessian_layout}~(\nameref{sec:hessian_layout}).

\begin{corollary}[Vectorized Kronecker-Product Representation of Fr\'echet Derivative]
\label{corollary:vectorized_kronecker_product_representation}
Let
\begin{align}
    f: \mathbb{R}^{m \times n} \to \mathbb{R}
\end{align}
be $\alpha$-times continuously Fr\'echet differentiable (\ie, $f\in C^\alpha$). For directions $H_1,\dots,H_\alpha\in\mathbb{R}^{m\times n}$, the multilinear action $D^{\alpha} f$ at $A \in \mathbb{R}^{m \times n}$ is given by the Frobenius tensor inner product \citep{kolda2009tensor,magnusmatrix}:
\begin{align}
D^\alpha f(A)[H_1,\dots,H_\alpha]
  &=
  \bigl\langle D^\alpha f(A), H_1\otimes\cdots\otimes H_\alpha \bigr\rangle \\
  &=\bigl\langle 
  \operatorname{vec}\left(D^\alpha f(A)\right),
  \operatorname{vec}\left(
  H_1\otimes\cdots\otimes H_\alpha \right)\bigr\rangle\\
  &=\operatorname{vec}\left(D^\alpha f(A)\right)^\top \operatorname{vec}\left(
  H_1\otimes\cdots\otimes H_\alpha 
  \right)
  ,
\end{align}
where $\otimes$ represents Kronecker product (\ie, tensor product) and:
\begin{align}
 \mathrm{vec}: \mathbb{R}^{m \times n} \mapsto \mathbb{R}^{mn}   
\end{align}
represents the vectorization operator with the \textbf{column-major convention} \citep{kolda2009tensor}. 
\end{corollary}

%\begin{remark}
This vectorization is particularly useful for representing arbitrary-order derivatives of matrix-valued functionals in matrix form.
%\end{remark}

\subsubsection{Representation Convention for Jacobian of Matrix-Valued Functional}
\label{sec:jacobian_layout}

Representing the Jacobian of matrix-valued functionals in matrix form is standard in the literature \citep{Horn2012}. For clarity, we introduce a matrix layout as a complementary representation for the Jacobian of matrix-valued functionals. Let
\begin{align}
f: \mathbb{R}^{m \times n} \mapsto \mathbb{R}  
\end{align}
be a first-order Fr\'echet differentiable functional. Then the differential of $f$ admits:
\begin{align}
    \dd f = D f(A)[\dd A]
  =\langle Df(A),\, \dd A\rangle  
  = \Bigl( \frac{\partial f}{\partial A} \Bigr)^\top \dd A
  =\tr\left[\left(\frac{\partial f}{\partial A}\right)^\top \dd A \right]
  ,
\end{align}
where $\frac{\partial f}{\partial A}$ and infinitesimal variation $\dd A \in \mathbb{R}^{m \times n}$ are piece-wisely defined as:
\begin{align}
    \frac{\partial f}{\partial A} = 
    \begin{bmatrix}
            \frac{\partial f}{\partial A_{1,1}} & \frac{\partial f}{\partial A_{1,2}} & \cdots & \frac{\partial f}{\partial A_{1,n}} \\
            \frac{\partial f}{\partial A_{2,1}} & \frac{\partial f}{\partial A_{2,2}} & \cdots & \frac{\partial f}{\partial A_{2,n}} \\
            \vdots & \vdots & \ddots & \vdots \\
            \frac{\partial f}{\partial A_{m,1}} & \frac{\partial f}{\partial A_{m,2}} & \cdots & \frac{\partial f}{\partial A_{m,n}}
        \end{bmatrix}
\end{align}
with \textbf{denominator layout convention}, and:
\begin{align}
    \dd A =\begin{bmatrix}
        \dd A_{1,1} & \dd A_{1,2} & \cdots & \dd A_{1,n} \\
        \dd A_{2,1} & \dd A_{2,2} & \cdots & \dd A_{2,n} \\
            \vdots & \vdots & \ddots & \vdots \\
        \dd A_{m,1} & \dd A_{m,2} & \cdots & \dd A_{m,n}
        \end{bmatrix}
        .
\end{align}

\subsubsection{Representation Convention for Hessian of Matrix-Valued Functional}
\label{sec:hessian_layout}

The Hessian of a matrix-valued functional is naturally a higher-order tensor; 
for instance, it is a fourth-order tensor for matrix-valued functionals \citep{kolda2009tensor}. Let
\begin{align}
    f: \mathbb{R}^{m \times n} \to \mathbb{R}
\end{align}
be a twice Fr\'echet differentiable functional. Since
\begin{align}
    D^2 f = D(Df),
\end{align}
to obtain a matrix representation of $D^2 f$, we first consider the representation layout of the first-order derivative for a matrix-to-matrix map $F: \mathbb{R}^{m \times n} \to \mathbb{R}^{s \times t}$. We then apply vectorization together with this layout to express the second-order derivatives of matrix-valued functionals in matrix form.

Let
\begin{align}
F: \mathbb{R}^{m \times n} \mapsto \mathbb{R}^{s \times t}    
\end{align}
be a first-order Fr\'echet differentiable matrix-to-matrix map. Then there exists:
\begin{align}
\frac{\partial \operatorname{vec}(F)}{\partial \operatorname{vec}(A)} \in \mathbb{R}^{p \times q}  \quad p=mn \quad \text{and} \quad q=st ,
\end{align}
piece-wisely defined as:
\begin{align}
    \frac{\partial \operatorname{vec}(F)}{\partial \operatorname{vec}(A)} &=
    \begin{pmatrix}
            \frac{\partial \operatorname{vec}(F)_{1}}{\partial \operatorname{vec}(A)_{1}} & \frac{\partial \operatorname{vec}(F)_{2}}{\partial \operatorname{vec}(A)_{1}} & \cdots & \frac{\partial \operatorname{vec}(F)_{q}}{\partial \operatorname{vec}(A)_{1}} \\
            \frac{\partial \operatorname{vec}(F)_{1}}{\partial \operatorname{vec}(A)_{2}} & \frac{\partial \operatorname{vec}(F)_{2}}{\partial \operatorname{vec}(A)_{2}} & \cdots & \frac{\partial \operatorname{vec}(F)_{q}}{\partial \operatorname{vec}(A)_{2}} \\
            \vdots & \vdots & \ddots & \vdots \\
            \frac{\partial \operatorname{vec}(F)_{1}}{\partial \operatorname{vec}(A)_{p}} & \frac{\partial \operatorname{vec}(F)_{2}}{\partial \operatorname{vec}(A)_{p}} & \cdots & \frac{\partial \operatorname{vec}(F)_{q}}{\partial \operatorname{vec}(A)_{p}}
        \end{pmatrix}
    \end{align}
by using \textbf{denominator layout convention} on $\operatorname{vec}(F)$ and $\operatorname{vec}(A)$ \citep{Horn2012}. Then the Hessian of the matrix-valued functional $f$ can be defined as:
\begin{align}
    \frac{\partial}{\partial \mathrm{vec}(A)} \mathrm{vec}\left( \frac{\partial f}{\partial A}\right)
\end{align}
with vectorized representation.

\bigskip
\noindent
\textbf{Relating Vectorized Representation to $D^2 f$.}
We now relate this vectorized representation to $D^2 f$. By Corollary~\ref{corollary:vectorized_kronecker_product_representation}~(\nameref{corollary:vectorized_kronecker_product_representation}), consider:
\begin{align}
D^2 f[\dd A, \dd A] 
    &= \langle D^2 f, \dd A \otimes \dd A \rangle
    ,
\end{align}
and use the following identities from Lemma~\ref{lemma:essential_matrix_identities}~(\nameref{lemma:essential_matrix_identities}):
\begin{enumerate}
    \item $\operatorname{vec}(x) = x$,
    \item $\operatorname{vec}(B V A^\top) = (A \otimes B)\operatorname{vec}(V)$,
\end{enumerate}
then it yields:
\begin{align}
D^2 f[\dd A, \dd A] 
&= \langle D^2 f, \dd A \otimes \dd A \rangle \\
&=   
\mathrm{vec}(D^2 f)^\top \mathrm{vec}(\dd A \otimes \dd A) \\
&= \mathrm{vec}(\dd A)^\top (D^2 f)^\top \mathrm{vec}(\dd A). 
\end{align}

\noindent
Relating 
\begin{align}
    \frac{\partial}{\partial \mathrm{vec}(A)} \mathrm{vec}\left( \frac{\partial f}{\partial A}\right)
\end{align}
with $D^2 f$ yields:
\begin{align}
    D^2f = \left[\frac{\partial}{\partial \mathrm{vec}(A)} \mathrm{vec}\left( \frac{\partial f}{\partial A}\right)\right]^\top
    ,
\end{align}
such that:
\begin{align}
    D^2 f[\dd A, \dd A] = \mathrm{vec}(\dd A)^\top 
    \left[
    \frac{\partial}{\partial \mathrm{vec}(A)} \mathrm{vec}\left( \frac{\partial f}{\partial A}\right)
    \right]
    \mathrm{vec}(\dd A).
\end{align}

\section{Refined Asymptotic Eigenvalue Expansion}
\label{sec:eigenvalue_expansion}

Kato's monograph \citep{kato1995perturbation} establishes the existence of asymptotic eigenvalue expansions and, in particular, provides a closed-form expression for the \emph{weighted mean} of eigenvalue coefficients, as stated in Theorem~\ref{thm:kato_weighted_mean}~(\nameref{thm:kato_weighted_mean}). Nevertheless, Kato's formulation is expressed with an infinite summation of contour integrals involving the perturbed resolvent and does not yield explicit, constructive formulas for the individual coefficients, which limits its direct applicability in our setting. Building on the analytic foundations laid by Kato, and by employing explicit Neumann expansions of resolvents together with the residue theorem, we refine this framework to derive an \emph{explicit, closed-form formula for arbitrary-order eigenvalue coefficients} of holomorphic families of bounded self-adjoint operators. Our main result, Theorem~\ref{thm:expansions_of_eigenvalues}~(\nameref{thm:expansions_of_eigenvalues}), goes beyond Kato's weighted mean by furnishing a fully constructive representation of each eigenvalue coefficient. The overall scheme is illustrated in Figure~\ref{fig:framework_of_spectral_analysis}.

\begin{definition}[Space of Bounded Linear Operators]
\label{def:bounded_linear_operators}
Let 
\begin{align}
 \mathcal L(X) = \{\,T : X \mapsto X \mid X \text{is a Banach space and}~T \text{ is a bounded linear operator}\}   
\end{align}
be the Banach space of bounded linear operators.     
\end{definition}

%page 77
\begin{theorem}[Kato's Weighted Mean of Eigenvalue Expansions {\citep[Ch.~II, {\S}2.2]{kato1995perturbation}}]
\label{thm:kato_weighted_mean}
Let 
\begin{align}
\mathcal{T}(x) = \mathcal{T}^{(0)} + \sum_{j=1}^\infty x^j \mathcal{T}^{(j)} \;\in\; \mathcal{L}(X)
\end{align}
be a holomorphic family of bounded operators on a Banach space $X$ \citep[Ch.~II, {\S}2.1, Eq~(2.1)]{kato1995perturbation}. Suppose $\lambda^{(0)}$ is an isolated eigenvalue of $\mathcal{T}^{(0)}$ of algebraic multiplicity $m$.

\medskip
\noindent
Let
\begin{align}
R(z) = (\mathcal{T}(x) - zI)^{-1}
\end{align}
be the perturbed resolvent, let
\begin{align}
  P(x) = -\frac{1}{2\pi i} \oint_{\Gamma} R(z) dz
  ,
\end{align}
be Riesz projector, and define
\begin{align}
\widehat{\mathcal{T}}^{(n)}
= \sum_{p=1}^\infty (-1)^{p-1}
\sum_{\substack{i_1 + \cdots + i_p = n \\ i_j \ge 1}}
\frac{1}{2\pi i}\oint_\Gamma
R(z)\,\mathcal{T}^{(i_1)}\,R(z)\cdots 
R(z)\,\mathcal{T}^{(i_p)}\,R(z)\,(z-\lambda^{(0)})\,dz
,
\end{align}
where $\Gamma$ is a small contour enclosing only $\lambda^{(0)}$ and no other spectrum \citep[Ch.~II, {\S}2.1, Eq~(2.18)]{kato1995perturbation}. Then the weighted mean of the perturbed eigenvalues is:
\begin{align}
\hat{\lambda}(x) := \frac{1}{m}\,\operatorname{tr}\!\bigl(\mathcal{T}(x) P(x)\bigr),
\end{align}
admits the expansion
\begin{align}
\hat{\lambda}(x) 
= \lambda^{(0)} + \sum_{n=1}^\infty x^n \hat{\lambda}^{(n)}
\end{align}
\citep[Ch.~II, {\S}2.1, Eq~(2.21)]{kato1995perturbation}, and:
\begin{align}
\hat{\lambda}^{(n)} = \tfrac{1}{m}\,\operatorname{tr}\bigl(\widehat{\mathcal{T}}^{(n)}\bigr)  
\end{align}
\citep[Ch.~II, {\S}2.1, Eq~(2.22)]{kato1995perturbation}.

\end{theorem}

\paragraph{\textbf{Sketch to Refine Kato's Result}.} 
Following Kato's analytic framework, we also begin with the perturbation series $\mathcal{T}(x)$ of a self-adjoint operator $\mathcal{T}$. Kato's monograph establishes eigenvalue expansions via contour integrals of the resolvent and provides a closed-form expression for the \emph{weighted mean} of eigenvalues, but it does not supply explicit constructive formulas for the individual coefficients. Our approach departs at this point: we expand the resolvent explicitly through its Neumann series, apply the residue theorem on the contour integrals, and simplify the resulting expressions. This yields a closed-form asymptotic eigenvalue expansion for $\mathcal{T}$ with computable coefficients. Unlike Kato's result, our expansion is formulated in terms of the finite summation of the series of unperturbed resolvent, which enables systematic computation and, in particular, facilitates the subsequent derivation of singular-value Fr\'echet derivatives. As a special case, it also recovers the classical Rayleigh--Schr\"odinger corrections of quantum mechanics \citep{rayleigh1896theory,schrodinger1926quantisierung,sakurai2020modern}.

\begin{theorem}[Refined Closed-Form Asymptotic Expansion of Simple Isolated Eigenvalue in Self-Adjoint Operator]
\label{thm:expansions_of_eigenvalues}
Let
\begin{align}
\mathcal{T}(x) &= \mathcal{T}^{(0)} + \sum_{j=1}^{\infty} x^j\,\mathcal{T}^{(j)} \;\in\; \mathcal{L}(X)
\end{align}
be a holomorphic family of bounded operators on a Banach space $X$, where $\mathcal{T}^{(0)}$ denotes the unperturbed operator and $x \in \mathbb{C}$ is the perturbation parameter.

\medskip
\noindent
\textbf{Unperturbed Reduced Resolvent.}
Define the unperturbed reduced resolvent \citep{Rudin1991,kato1995perturbation,yosida2012functional} of $\mathcal{T}^{(0)}$:
\begin{align}
S^{(0)} &= \bigl(\mathcal{T}^{(0)} - \lambda^{(0)} I\bigr)^{-1}\,(I - w^{(0)}(w^{(0)})^*)
,
\end{align}
where $\lambda^{(0)}$ is a simple eigenvalue of $\mathcal{T}^{(0)}$ and $w^{(0)}$ is the associated normalized eigenvector (\ie, $\|w^{(0)}\|_2=1$). 

\medskip
\noindent
\textbf{Theorem Claim.}
Then there exists a unique holomorphic branch $\lambda(x)$ of eigenvalues of $\mathcal{T}(x)$. It admits the power series:
\begin{align}
\lambda(x) &= \sum_{n=0}^{\infty} x^n\,\lambda^{(n)}, 
\end{align}
and for each integer $n\ge1$,
\begin{align}
\lambda^{(n)}
&=
\sum_{p=1}^n(-1)^{\,p-1}
\sum_{\substack{i_1+\cdots+i_p=n\\i_j\ge1}}
\bigl\langle w^{(0)},\,
  \mathcal{T}^{(i_1)}\,S^{(0)}\,\mathcal{T}^{(i_2)}\,S^{(0)}\cdots S^{(0)}\,\mathcal{T}^{(i_p)}\,w^{(0)}
\bigr\rangle.
\end{align}
\end{theorem}

\begin{mynewproof}

We begin by presenting a compact and explicit proof of Theorem~\ref{thm:kato_weighted_mean}~(\nameref{thm:kato_weighted_mean}) in the case of a simple eigenvalue of a self-adjoint operator, which does not exist in Kato's monograph. Since $\lambda^{(0)}$ is a simple eigenvalue of $\mathcal{T}^{(0)}$, analytic‐perturbation theory ensures there exists a unique eigenvalue branch $\lambda(x)$. Let
\begin{align}
    R(z) = (\mathcal{T}(x) - zI)^{-1}
\end{align}
be the \emph{resolvent} of operator $\mathcal{T}(x)$, which encapsulates the full spectral information of $\mathcal{T}(x)$, and let
\begin{align}
    S(z) = R(z)(I - P(x))
\end{align}
be the associated reduced resolvent $S(z)$ (\ie, the regular part of the resolvent), where $P(x)$ is the Riesz--Dunford contour integral \citep{dunford1988linear,kato1995perturbation}, that is:
\begin{align}
P(x) =
 -\frac{1}{2\pi i}\oint_{\Gamma} R(z) \,dz
=
 -\frac{1}{2\pi i}\oint_{\Gamma} \bigl(\mathcal{T}(x) - zI\bigr)^{-1}\,dz \label{equ:dunford_riesz_contour_integral}
 , 
\end{align}
where $\Gamma$ is a small contour enclosing only $\lambda^{(0)}$ and no other eigenvalues. Hence the projection:
\begin{align}
\mathcal{T}(x)\,P(x)
=-\frac{1}{2\pi i}\oint_{\Gamma} \bigl[I + zR(z) \bigr] dz
=\frac{1}{2\pi i}\oint_{\Gamma} -zR(z)  dz
=
\lambda(x)\,P(x) \label{equ:identity_Riesz_projector}
,
\end{align}
holds true since the resolvent $R(z)$ for a simple eigenvalue admits the Laurent expansion \citep[Ch.~I, {\S}5.3, Eq~(5.18)]{kato1995perturbation}
\begin{align}
R(z) = \frac{P(x)}{\lambda(x)-z} + S(z) + \hat{R}(z) 
, \quad
\hat{R}(z) = \sum_{n=1}^{\infty} \big(z - \lambda(x)\big)^{n} \big(S(z)\big)^{n+1}
\label{equ:laurent_expansion_resolvent}
,    
\end{align}
where $\hat{R}(z)$ is the analytic regular remainder of $z$, and the residue of $-zR(z)$ at $z=\lambda(x)$ is
\begin{align}
\mathrm{Res}_{z=\lambda(x)}(-zR(z))= [z - \lambda(x)] [-zR(z)]\mid_{z=\lambda(x)}=
 \lambda(x) P(x)
.
\end{align}

\bigskip
\noindent
\textbf{Contour Integral of Perturbed Eigenvalue Series $\lambda(x)$.} 
Starting from the identity in Equation~(\ref{equ:identity_Riesz_projector}),
\begin{align}
\mathcal{T}(x)P(x) &= \lambda(x)P(x) \\
\Longrightarrow 
\tr\bigl(\mathcal{T}(x)P(x)\bigr) &= \tr\bigl(\lambda(x)P(x)\bigr) \\
&= \tr\bigl(\lambda(x)\bigr)\, \tr\bigl(P(x)\bigr) \\
&= \lambda(x)\\
\Longrightarrow 
   \lambda(x) - \lambda^{(0)} &=  \tr\bigl(\mathcal{T}(x)P(x)\bigr) - \lambda^{(0)} \\
   &= \tr\bigl(\mathcal{T}(x)P(x)\bigr) - \tr\bigl(\lambda^{(0)} P(x) \bigr) \\
   &=\tr\Bigl(\bigl(\mathcal{T}(x) -\lambda^{(0)}I\bigr) P(x)\Bigr) 
   ,
\end{align}
then substituting $P(x)$ from Equation~(\ref{equ:identity_Riesz_projector}) yields:
\begin{align}
\lambda(x) - \lambda^{(0)} &=    \tr\Bigl(\bigl(\mathcal{T}(x) -\lambda^{(0)}I\bigr) \bigl( -\frac{1}{2\pi i}\oint_{\Gamma} R(z) \,dz \bigr) \Bigr) \\
&=-\frac{1}{2\pi i}\oint_\Gamma
\operatorname{tr}\!\big((T(x)-\lambda^{(0)}I)R(z)\big)\,dz 
.
\end{align}
Considering the resolvent identity:
\begin{align}
    (\mathcal T(x)-zI)R(z) = I
\end{align}
and:
\begin{align}
\bigl(\mathcal T(x)-\lambda^{(0)} I\bigr)R(z) 
&=\Bigl(\mathcal T(x)-zI + zI-\lambda^{(0)}I\Bigr)R(z) \\
&= \Bigl(\mathcal T(x)-zI + (z-\lambda^{(0)})I\Bigr)R(z) \\
&=(\mathcal T(x)-zI)R(z) + (z-\lambda^{(0)})R(z) \\
&=I + (z-\lambda^{(0)})R(z)
,
\end{align}
it yields:
\begin{align}
\lambda(x) - \lambda^{(0)} 
&=
-\frac{1}{2\pi i}\oint_\Gamma
\operatorname{tr}\!\big((T(x)-\lambda^{(0)}I)R(z)\big)\,dz \\
&=-\frac{1}{2\pi i}\oint_\Gamma
\operatorname{tr}\!\big(
I + (z-\lambda^{(0)})R(z)
\big)\,dz \\
&=-\frac{1}{2\pi i}\oint_\Gamma
\operatorname{tr}\!\big((z-\lambda^{(0)})R(z)
\big)\,dz \\
&=-\frac{1}{2\pi i}\oint_\Gamma
 \operatorname{tr}\!\big(
 (z-\lambda^{(0)}) R(z)
 \big) \operatorname{tr}\!\big(1\big)\,dz \\
&=-\frac{1}{2\pi i}\oint_\Gamma
 \operatorname{tr}\!\big(
 (z-\lambda^{(0)})R(z)
 \big) \operatorname{tr}\!\big(P^{(0)}\big)\,dz \\
&=-\frac{1}{2\pi i}\oint_\Gamma
 \operatorname{tr}\!\big(
 (z-\lambda^{(0)}) R(z) P^{(0)}
 \big) \,dz 
\label{equ:contour_integral_lambdax_lambda0}
.
\end{align}

\bigskip
\noindent
\textbf{Resolvent Expansion.}
To expand the resolvent $R(z)$, define the unperturbed resolvent $R^{(0)}(z)$: 
\begin{align}
R^{(0)}(z) = \bigl(\mathcal{T}^{(0)} - zI\bigr)^{-1}
,
\end{align}
then this identity holds:
\begin{align}
 (\mathcal{T}^{(0)} - zI) R^{(0)}(z) = I .   
\end{align}

\noindent
Note that:
\begin{align}
\mathcal{T}(x) - zI
&= \mathcal{T}^{(0)} - zI + \sum_{j=1}^\infty x^j \mathcal{T}^{(j)} \\
&=\mathcal{T}^{(0)} - zI + I\sum_{j=1}^\infty x^j \mathcal{T}^{(j)} \\
&=\mathcal{T}^{(0)} - zI + (\mathcal{T}^{(0)} - zI) R^{(0)}(z)\sum_{j=1}^\infty x^j \mathcal{T}^{(j)} \\
&= (\mathcal{T}^{(0)} - zI)\Bigl(I + R^{(0)}(z)\,\sum_{j=1}^\infty x^j \mathcal{T}^{(j)}\Bigr)
,
\end{align}
so that the operator Neumann series of $R(z)$ expands as:
\begin{align}
R(z)&=\bigl(\mathcal{T}(x)-zI\bigr)^{-1} \\
&= \left[ (\mathcal{T}^{(0)} - zI)\Bigl(I + R^{(0)}(z)\,\sum_{j=1}^\infty x^j \mathcal{T}^{(j)}\Bigr) \right]^{-1}\\
&= \Bigl(I + R^{(0)}(z)\,\sum_{j=1}^\infty x^j \mathcal{T}^{(j)}\Bigr)^{-1} \Bigl(\mathcal{T}^{(0)} - zI\Bigr)^{-1}\\
&= \Bigl(I + R^{(0)}(z)\,\sum_{j=1}^\infty x^j \mathcal{T}^{(j)}\Bigr)^{-1} R^{(0)}(z)\\
&= \underbrace{
\Bigl(I - \bigl(-R^{(0)}(z)\,\sum_{j=1}^\infty x^j \mathcal{T}^{(j)}\bigr) \Bigr)^{-1}
}_{\text{Neumann series}} R^{(0)}(z) \\
&=\sum_{k=0}^\infty 
(-1)^k \Bigl(R^{(0)}(z)\,\sum_{j=1}^\infty x^j\mathcal{T}^{(j)}\Bigr)^k R^{(0)}(z)
\label{equ:Tx_minus_zI_2}
.
\end{align}

\bigskip
\noindent
\textbf{Asymptotic Eigenvalue Expansion.}
Expanding the term in Equation~(\ref{equ:Tx_minus_zI_2}):
\begin{align}
\Bigl(R^{(0)}(z)\,\sum_{j=1}^\infty x^j\mathcal{T}^{(j)}\Bigr)^k 
\end{align}
yields:
\begin{align}
\Bigl(R^{(0)}(z)\,\sum_{j=1}^\infty x^j\mathcal{T}^{(j)}\Bigr)^k &=   \sum_{i_1,\dots,i_k\ge1}
x^{\,i_1+\cdots+i_k}\,
R^{(0)}(z)\,\mathcal{T}^{(i_1)}\,R^{(0)}(z)\cdots R^{(0)}(z)\,\mathcal{T}^{(i_k)}    \\
&=\sum_{i_1,\dots,i_k\ge1}
x^{\,i_1+\cdots+i_k}\, R^{(0)}(z)\, R_{i_k}(z) \label{equ:Tx_minus_zI_3}
,
\end{align}
where $R_{i_k}(z)$ represents an operator composition series:
\begin{align}
    R_{i_k}(z)= \mathcal{T}^{(i_1)}\,R^{(0)}(z)\cdots R^{(0)}(z)\,\mathcal{T}^{(i_k)}
    .
\end{align}
Substituting Equation~(\ref{equ:Tx_minus_zI_3})
into Equation~(\ref{equ:Tx_minus_zI_2}) yields:
\begin{align}
R(z)
&= \sum_{k=0}^\infty(-1)^k
\sum_{i_1,\dots,i_k\ge1}
x^{\,i_1+\cdots+i_k}\,
R^{(0)}(z)R_{i_k}(z)R^{(0)}(z)  \\
&= \sum_{k=0}^\infty(-1)^k
\sum_{i_1,\dots,i_k\ge1}
x^{\,i_1+\cdots+i_k}\,
R^{(0)}(z)R_{i_k}(z)R^{(0)}(z)
\label{equ:expansion_Tx_minus_zI} 
.
\end{align}
Substituting Equation~(\ref{equ:expansion_Tx_minus_zI}) into the contour integral for $\lambda(x)-\lambda^{(0)}$ in Equation~(\ref{equ:contour_integral_lambdax_lambda0}):
\begin{align}
\lambda(x)-\lambda^{(0)} =  -\frac{1}{2\pi i}\oint_\Gamma
 \operatorname{tr}\!\big((z-\lambda^{(0)}) R(z) P^{(0)}
 \big) \,dz 
  ,
\end{align}
yields:
\begin{align}
\lambda^{(n)}
&=\sum_{k=0}^\infty(-1)^k
\!\!\!\sum_{\substack{i_1+\cdots+i_k=n\\i_j\ge1}}
-\frac{1}{2\pi i}
\oint_\Gamma
\tr\Bigl( 
(z-\lambda^{(0)}) R^{(0)}(z) R_{i_k}(z)R^{(0)}(z) P^{(0)}
\Bigr)
dz \\
&=\sum_{k=0}^{\infty} (-1)^{k-1}
\!\!\!\sum_{\substack{i_1+\cdots+i_k=n\\i_j\ge1}}
\frac{1}{2\pi i}
\oint_\Gamma
\tr\Bigl( 
(z-\lambda^{(0)}) R^{(0)}(z) R_{i_k}(z) R^{(0)}(z) P^{(0)}
\Bigr)
dz 
,
\end{align}
which recovers Theorem~\ref{thm:kato_weighted_mean}~(\nameref{thm:kato_weighted_mean}).

\bigskip
\noindent
\textbf{Contracting and Relabeling Indices.} We refine Theorem~\ref{thm:kato_weighted_mean}~(\nameref{thm:kato_weighted_mean}) further, with the aim of obtaining a constructive, computable and basis-dependent formulation. Note that only the multi-indices satisfying:
\begin{align}
    i_1 + \cdots + i_k = n    
\end{align}
contribute to the coefficient of $x^n$, and since the terms with:
\begin{align}
    k = 0 \quad \text{or} \quad k > n
\end{align}
for $n \geq 1$ vanish, we contract the summation to the admissible subset of indices. For clarity, we denote this contracted index set by $p \subseteq k$:
\begin{align}
\lambda^{(n)}
    &=\sum_{p=1}^{n} (-1)^{p-1}
\!\!\!\sum_{\substack{i_1+\cdots+i_p=n\\i_j\ge1}}
\frac{1}{2\pi i}
\oint_\Gamma
\tr\Bigl(
(z-\lambda^{(0)}) R^{(0)}(z) R_{i_p}(z)R^{(0)}(z)\,P^{(0)}
\Bigr)
dz  
\label{equ:pre_residue_eigenvalue_expansion}
.
\end{align}

\bigskip
\noindent
\textbf{Applying Cauchy's Residue Theorem}
Applying Cauchy's residue theorem via Riesz--Dunford functional calculus \citep{dunford1988linear} for Equation~(\ref{equ:pre_residue_eigenvalue_expansion}) yields:
\begin{align}
\lambda^{(n)}
&=\sum_{p=1}^n(-1)^{p-1}
\sum_{\substack{i_1+\cdots+i_p=n\\i_j\ge1}}
\operatorname{Res}_{z=\lambda^{(0)}} \Bigl[
\tr\Bigl(
(z-\lambda^{(0)}) 
R^{(0)}(z) R_{i_p}(z) R^{(0)}(z) P^{(0)} 
\Bigr)
\Bigr]
\label{equ:expansion_lambda_n}
.
\end{align}

\bigskip
\noindent
\textbf{Simplifying Residue.} We now aim to compute the residue:
\begin{align}
\operatorname{Res}_{z=\lambda^{(0)}} \Bigl[
\tr \Bigl(
(z - \lambda^{(0)}) R^{(0)}(z) R_{i_p}(z) R^{(0)}(z) P^{(0)} 
\Bigr)
\Bigr]
.
\end{align}
Note that near $z=\lambda^{(0)}$, the Laurent expansion of the unperturbed resolvent $R^{(0)}(z)$ for a simple eigenvalue $\lambda^{(0)}$ admits \citep[Ch.~I, {\S}5.3, Eq~(5.18)]{kato1995perturbation}:
\begin{align}
R^{(0)}(z) 
=\frac{P^{(0)}}{\lambda^{(0)}-z} 
+ S^{(0)} + \hat{R}^{(0)}(z),
\qquad
\hat{R}^{(0)}(z) = \sum_{n=1}^{\infty} \big(z - \lambda^{(0)}\big)^{n} \big(S^{(0)}\big)^{n+1}
,
\end{align}
where $P^{(0)} = w^{(0)}(w^{(0)})^*$, $S^{(0)}=R^{(0)}(z)(I-P^{(0)})$ is the unperturbed reduced resolvent of $R^{(0)}(z)$, and $\hat{R}^{(0)}(z)$ is the analytic regular remainder of $z$. Substitute $R^{(0)}(z)$ into the trace product:
\begin{align}
\tr\Bigl(
(z - \lambda^{(0)}) R^{(0)}(z) R_{i_p}(z) R^{(0)}(z) P^{(0)}
\Bigr)
,    
\end{align}
and consider that in the expanded trace product:
\begin{enumerate}
    \item the terms $P^{(0)}S^{(0)}=S^{(0)}P^{(0)}=0$ vanish,

    \item the terms with higher-order poles vanish, since the denominators are constant operators.
    
\end{enumerate}
Then only the term with simple pole survives:\\
\pgfmathsetmacro{\linewidthfactor}{0.9}
\pgfmathsetmacro{\inverselinewidthfactor}{1/\linewidthfactor}
\scalebox{\linewidthfactor}{% scale
\begin{minipage}{\inverselinewidthfactor\linewidth} % 1/scale
\begin{align}
\operatorname{Res}_{z=\lambda^{(0)}} \Bigl[
\tr\Bigl(
(z - \lambda^{(0)}) R^{(0)}(z) R_{i_p}(z) R^{(0)}(z) P^{(0)}
\Bigr)
\Bigr]
&=
\operatorname{Res}_{z=\lambda^{(0)}} \Bigl[
\tr\Bigl(
(z - \lambda^{(0)})
\frac{P^{(0)}\mathcal T^{(i_1)}S^{(0)}\cdots S^{(0)}\mathcal T^{(i_p)}P^{(0)}}{(\lambda^{(0)}-z)^2} P^{(0)}
\Bigr)
\Bigr] \\
&=\operatorname{Res}_{z=\lambda^{(0)}} \Bigl[
\tr\Bigl(
\frac{P^{(0)}\mathcal T^{(i_1)}S^{(0)}\cdots S^{(0)}\mathcal T^{(i_p)}P^{(0)}}{z-\lambda^{(0)}}  P^{(0)}
\Bigr)
\Bigr] \\
&=(z-\lambda^{(0)}) 
\tr\Bigl(
\frac{P^{(0)}\mathcal T^{(i_1)}S^{(0)}\cdots S^{(0)}\mathcal T^{(i_p)}P^{(0)}}{z-\lambda^{(0)}} P^{(0)} 
\Bigr)
\mid_{z \to \lambda^{(0)}} 
\\
&=
\tr\Bigl(
P^{(0)}\mathcal T^{(i_1)}S^{(0)}\cdots S^{(0)}\mathcal T^{(i_p)}P^{(0)}  P^{(0)}
\Bigr) \\
&=
\tr\Bigl(
\mathcal T^{(i_1)}S^{(0)}\cdots S^{(0)}\mathcal T^{(i_p)}P^{(0)}  P^{(0)} P^{(0)}
\Bigr) \\
&=
\tr\Bigl(
\mathcal T^{(i_1)}S^{(0)}\cdots S^{(0)}\mathcal T^{(i_p)} P^{(0)}  
\Bigr)
\label{equ:residue_of_inner_product}
,
\end{align}
\end{minipage}
}\\
since $P^{(0)}P^{(0)}=P^{(0)}$.

\bigskip
\noindent
\textbf{Producing Theorem Claim.}
Substituting the residue from Equation~(\ref{equ:residue_of_inner_product}) into Equation~(\ref{equ:expansion_lambda_n}) yields:
\begin{align}
\lambda^{(n)} &= 
\sum_{p=1}^n (-1)^{p-1} \sum_{\substack{i_1 + \cdots + i_p = n \\ i_j \geq 1}} 
\tr\Bigl(
\mathcal T^{(i_1)}S^{(0)}\cdots S^{(0)}\mathcal T^{(i_p)} P^{(0)}  
\Bigr) \\
&= 
\sum_{p=1}^n (-1)^{p-1} \sum_{\substack{i_1 + \cdots + i_p = n \\ i_j \geq 1}} 
\tr\Bigl(
\mathcal T^{(i_1)}S^{(0)}\cdots S^{(0)}\mathcal T^{(i_p)} w^{(0)} (w^{(0)})^*  
\Bigr) \\
&= 
\sum_{p=1}^n (-1)^{p-1} \sum_{\substack{i_1 + \cdots + i_p = n \\ i_j \geq 1}} 
\tr\Bigl(
\bigl(w^{(0)}\bigr)^*  
\mathcal T^{(i_1)}S^{(0)}\cdots S^{(0)}\mathcal T^{(i_p)} w^{(0)} 
\Bigr) \\
&=\sum_{p=1}^n (-1)^{p-1} \sum_{\substack{i_1 + \cdots + i_p = n \\ i_j \geq 1}} 
\Bigl\langle w^{(0)}, \mathcal{T}^{(i_1)} S^{(0)} \mathcal{T}^{(i_2)} S^{(0)} \cdots S^{(0)} \mathcal{T}^{(i_p)} w^{(0)} \Bigr\rangle 
,
\end{align}
which is basis-dependent and expressed in terms of unperturbed reduced resolvent.

{\hfill$\qed$\par}
\end{mynewproof}

\begin{remark}
By refining Theorem~\ref{thm:kato_weighted_mean}~(\nameref{thm:kato_weighted_mean}), Theorem~\ref{thm:expansions_of_eigenvalues}~(\nameref{thm:expansions_of_eigenvalues}) provides an \emph{explicit, closed-form representation} of the coefficients $\lambda^{(n)}$ in the eigenvalue perturbation series. Classical analytic perturbation theory \citep{kato1995perturbation} guarantees the existence of such expansions and gives recursive characterizations of the coefficients, but does not furnish constructive closed forms. In contrast, our formulation expresses each $\lambda^{(n)}$ in terms of finitely many operator products involving the perturbation operators $\mathcal{T}^{(j)}$ and the unperturbed reduced resolvent $S^{(0)}$, making the coefficients directly computable. As a validation, for $n=1,2,\dots$, the expansion specializes to the familiar Rayleigh--Schr\"odinger corrections of quantum mechanics \citep{rayleigh1896theory,schrodinger1926quantisierung,sakurai2020modern}.
\end{remark}

\section{Infinitesimal Higher-Order Spectral Variations}
\label{sec:spectral_variation}

Guided by the scheme in Figure~\ref{fig:framework_of_spectral_analysis}, and under Assumption~\ref{assump:simple_singular_values}~(\nameref{assump:simple_singular_values}), we exploit the spectral correspondence between a rectangular matrix $A \in \mathbb{R}^{m \times n}$ and its Jordan--Wielandt embedding $\mathcal{T}$ as established in Theorem~\ref{theorem:Jordan_Wielandt_relation}~(\nameref{theorem:Jordan_Wielandt_relation}). This allows us to derive arbitrary--order Fr\'echet derivatives of the singular values of $A$ from the asymptotic eigenvalue expansions of $\mathcal{T}$. The argument proceeds by first establishing the correspondence between the perturbation series and Fr\'echet derivatives as stated in Theorem~\ref{thm:frechet_taylor_operator}~(\nameref{thm:frechet_taylor_operator}), and then applying this relation to obtain higher--order derivatives of singular values as stated in Theorem~\ref{thm:nth_order_singular_value}~(\nameref{thm:nth_order_singular_value}).

%A holomorphic family of type (A) admits a power series expansion in x with coefficients given by the Fréchet derivatives.
\begin{theorem}[Analytic Perturbation for Holomorphic Operators]
\label{thm:frechet_taylor_operator}
Let $X$ be a Banach space and let $\mathcal{T}(x):U\subset\mathbb{C}\to\mathcal{L}(X)$ be a holomorphic family --- \ie, type (A) in the sense of Kato's framework \citep{kato1995perturbation}, defined in a neighborhood of $0$ in the operator norm. Then $\mathcal{T}$ is $C^\infty$ in the Fr\'echet sense at $0$ and admits the convergent operator--norm expansion:
\begin{align}
\mathcal{T}(x) \;=\; \sum_{n=0}^{\infty}\frac{x^{n}}{n!}\,D^{n}\mathcal{T}(0),
\qquad |x|<\rho,
\end{align}
where $\rho$ is the distance from $0$ to the boundary of $U$. In particular, if one writes the perturbation series as:
\begin{align}
\mathcal{T}(x) = \mathcal{T}^{(0)}+\sum_{n=1}^{\infty}x^{n}\,\mathcal{T}^{(n)},
\end{align}
then the coefficients agree with the Fr\'echet derivatives, namely:
\begin{align}
\mathcal{T}^{(n)} = \tfrac{1}{n!}\,D^{n}\mathcal{T}(0), \qquad n\ge1.
\end{align}

\end{theorem}

\begin{theorem}[Higher-Order Infinitesimal Spectral Variation]
\label{thm:nth_order_singular_value}

Let 
\begin{align}
A \in \mathbb{R}^{m \times n}    
\end{align}
be a real rectangular matrix under Assumption~\ref{assump:simple_singular_values}~(\nameref{assump:simple_singular_values}).

\bigskip
\noindent
\textbf{Matrix Perturbation Series.} 
Let 
\begin{align}
A(x)=\sum_{k=0}^\infty x^k\,A^{(k)}\in\mathbb{R}^{m\times n} 
\end{align}
be holomorphic perturbed operator near $x=0$ with unperturbed matrix $A^{(0)} = A$. 

The unperturbed matrix $A^{(0)}$ admits a full SVD:
\begin{align}
A^{(0)} = U^{(0)} \Sigma^{(0)} (V^{(0)})^T,    
\end{align}
as defined in Theorem~\ref{theorem:full_svd}~(\nameref{theorem:full_svd}), where ordered $r=\rank(A)$ non-zero singular values are given, under Assumption~\ref{assump:simple_singular_values}~(\nameref{assump:simple_singular_values}), as:
\begin{align}
\sigma_1^{(0)} > \sigma_2^{(0)} > \ldots > \sigma_r^{(0)} > 0
,
\end{align}
$u_k^{(0)}$ and $v_k^{(0)}$ are singular vectors associated with singular value $\sigma_k^{(0)}$, and:
\begin{align}
   U^{(0)} \in \mathbb{R}^{m \times m} , \quad  V^{(0)} \in \mathbb{R}^{n \times n}
\end{align}
are orthogonal matrices. For brevity, we also use $\sigma_i = \sigma_i^{(0)}$, $u_i=u_i^{(0)}$, and $v_i=v_i^{(0)}$.

\bigskip
\noindent
\textbf{Jordan--Wielandt Perturbation Series Embedding.} 
Using Theorem~\ref{theorem:Jordan_Wielandt_relation}~(\nameref{theorem:Jordan_Wielandt_relation}), we embed the perturbation series $A(x)$ into $\mathcal{T}(x)$ to construct a Jordan--Wielandt embedding:
\begin{align}
\mathcal{T}(x) = \begin{bmatrix}
0 & A(x) \\
A(x)^\top & 0
\end{bmatrix}
.
\end{align}
This embedding admits a perturbation series:
\begin{align}
\mathcal{T}(x) = \sum_{j=0}^\infty x^j \mathcal{T}^{(j)}
\end{align}
at $x$ near zero, with the unperturbed operator:
\begin{align}
\mathcal{T}^{(0)} = 
\begin{bmatrix} 
0 & A^{(0)} \\ 
(A^{(0)})^\top & 0 
\end{bmatrix},    
\end{align}
and the perturbations:
\begin{align}
\mathcal{T}^{(j)} = \begin{bmatrix} 0 & A^{(j)} \\ (A^{(j)})^\top & 0 \end{bmatrix}, \quad j \geq 1.    
\end{align}

The non-zero eigenvalues $\lambda_i^{(\pm 0)}$ of $\mathcal{T}^{(0)}$ are therefore:
\begin{align}
 \lambda_i^{(+0)} = + \sigma_i^{(0)},
 \quad 
 \lambda_i^{(-0)} = - \sigma_i^{(0)},
 \quad \text{for}~i = 1, \ldots, r,   
\end{align}
associated with eigenvectors:
\begin{align}
w_i^{(+0)} = \frac{1}{\sqrt{2}} 
\begin{bmatrix} 
u_i^{(0)} \\ 
v_i^{(0)} 
\end{bmatrix}
, 
\quad w_i^{(-0)} = \frac{1}{\sqrt{2}} 
\begin{bmatrix} 
u_i^{(0)} \\ 
-v_i^{(0)} 
\end{bmatrix}
,    
\end{align}
and null eigenvectors, 
\begin{align}
a_j^{(0)} = 
\begin{bmatrix} 
u_j^{(0)} \\ 
0 
\end{bmatrix}  
, \quad (\text{for}~ j=r+1,\cdots,m)    ,
\end{align}
and:
\begin{align}
b_k^{(0)}=
\begin{bmatrix} 
0 
\\ 
v_k^{(0)} 
\end{bmatrix}
, \quad (\text{for}~ j=r+1,\cdots, n) .   
\end{align}

Since $w_k^{(+0)}$ and $w_k^{(-0)}$ are eigenvalues of $T^{(0)}$, hence the identities hold:
\begin{align}
\mathcal{T}^{(0)} w_k^{(+0)} = \lambda_k^{(+0)} w_k^{(+0)}
\Longrightarrow
\mathcal{T}^{(0)} w_k^{(+0)} = \sigma_k^{(0)} w_k^{(+0)}, 
\end{align}
and:
\begin{align}
\mathcal{T}^{(0)} w_k^{(-0)} =   \lambda_k^{(-0)} w_k^{(-0)}
\Longrightarrow
\mathcal{T}^{(0)} w_k^{(-0)} =   -\sigma_k^{(0)} w_k^{(-0)}.  
\end{align}

\bigskip
\noindent
\textbf{Unperturbed Reduced Resolvent in Embedding.} 
By definition, the spectral expansion of the reduced resolvent operator associated with the eigenvalue $\lambda_k^{(+0)}=\sigma_k^{(0)}$ and associated eigenvector $w_k^{(+0)}$ of $\mathcal{T}(x)$ is given as:
\begin{align}
S_k^{(0)} = \left( \mathcal{T}^{(0)} - \sigma_k^{(0)} I \right)^{-1} \left( I - P_k^{(0)} \right), \quad P_k^{(0)} = w_k^{(+0)} (w_k^{(+0)})^\top
,
\end{align}
which admits the spectral expansion:\\
\pgfmathsetmacro{\linewidthfactor}{0.9}
\pgfmathsetmacro{\inverselinewidthfactor}{1/\linewidthfactor}
\scalebox{\linewidthfactor}{% scale
\begin{minipage}{\inverselinewidthfactor\linewidth} % 1/scale
\begin{align}
S_k^{(0)} = 
\sum_{i=1, ~i \neq k}^{r} \frac{w_i^{(+0)} (w_i^{(+0)})^\top}{\sigma_i^{(0)} - \sigma_k^{(0)}} + 
\sum_{i=1,i \neq k}^r \frac{w_i^{(-0)} (w_i^{(-0)})^\top}{-\sigma_i^{(0)} - \sigma_k^{(0)}} - 
\sum_{j=r+1}^m \frac{a_j^{(0)} (a_j^{(0)})^\top}{\sigma_k^{(0)}} - \sum_{j=r+1}^n \frac{b_j^{(0)} (b_j^{(0)})^\top}{\sigma_k^{(0)}}.
\end{align}
\end{minipage}
}

\bigskip
\noindent
\textbf{Theorem Claim.} 
For each integer $n \ge 1$,
\begin{align}
\sigma_k^{(n)} = \sum_{p=1}^n (-1)^{p-1} \sum_{\substack{i_1 + \cdots + i_p = n \\ i_j \geq 1}} \langle w_k^{(+0)}, \mathcal{T}^{(i_1)} S_k^{(0)} \mathcal{T}^{(i_2)} \cdots S_k^{(0)} \mathcal{T}^{(i_p)} w_k^{(+0)} \rangle.
\end{align}
By Theorem~\ref{thm:frechet_taylor_operator}~(\nameref{thm:frechet_taylor_operator}), the Fr\'echet derivative of the singular value is unique and given by:
\begin{align}
D^n \sigma_k [\dd A, \cdots, \dd A] = n! \lim_{x \to 0} \left(x^n \; \sigma_k^{(n)}\right)
,
\end{align}
where:
\begin{align}
\dd A = \lim_{x \to 0} x A^{(1)}  .
\end{align}

\end{theorem}

\begin{remark}[\textbf{Schematic Procedure of Computing Higher-Order Singular-Value Derivatives}]
\label{remark:schematic_procedure_singular_value_derivative}
The suggested schematic procedure of computing arbitrary higher-order singular-value derivatives is as follows:
\begin{itemize}
    \item \textbf{Procedure I -- Construct Infinitesimal Perturbation} --- constructs an infinitesimal perturbation by $\dd A = \lim_{x \to 0} xA^{(1)}$,
    
    \item \textbf{Procedure II -- Specialize $n$ to Obtain Derivative Operator} --- specialize $n$ in $\sigma^{(n)}$, and obtain derivative operator $D^n\sigma_k = n! \sigma^{(n)}$,
    
    \item \textbf{Procedure III -- Map Derivative Operator Layout} --- map $D^n\sigma_k[\dd A,\cdots, \dd A]$ to Kronecker-product representation or specific layout.
\end{itemize}
\end{remark}

\begin{mynewproof}

By Theorem~\ref{thm:expansions_of_eigenvalues}~(\nameref{thm:expansions_of_eigenvalues}), one eigenvalue $\lambda_k(x)$ of $\mathcal{T}(x)$ admits an asymptotic expansion:
\begin{align}
\lambda_k(x) = \sum_{n=0}^\infty x^n \lambda_k^{(n)}    
.
\end{align}

By Theorem~\ref{theorem:Jordan_Wielandt_relation}~(\nameref{theorem:Jordan_Wielandt_relation}), for $n \geq 1$, choosing a positive eigenvalue branch $\sigma_k^{(0)}$ yields the asymptotic singular-value expansion of $A(x)$:
\begin{align}
\sigma_k^{(n)} = \sum_{p=1}^n (-1)^{p-1} \sum_{\substack{i_1 + \cdots + i_p = n \\ i_j \geq 1}} \langle w_k^{(+0)}, \mathcal{T}^{(i_1)} S_k^{(0)} \mathcal{T}^{(i_2)} \cdots S_k^{(0)} \mathcal{T}^{(i_p)} w_k^{(+0)} \rangle.    
\end{align}

By Theorem~\ref{thm:frechet_taylor_operator}~(\nameref{thm:frechet_taylor_operator}), $D^n \sigma_k$ admits:
\begin{align}
D^n \sigma_k = n! \sigma_k^{(n)}
,
\end{align}
and its action is given by:
\begin{align}
D^n \sigma_k[\dd A, \cdots, \dd A]   = n! \lim_{x \to 0} x^n \sigma_k^{(n)}
,
\end{align}
where
\begin{align}
    \dd A = \lim_{x \to 0} x A^{(1)}
    .
\end{align}

{\hfill$\qed$\par}
\end{mynewproof}

\begin{remark}
\textbf{Thanks to Kato's perturbation theory for linear operators, our framework for deriving singular-value derivatives rests on a rigorous analytic foundation and provides a procedural and systematic methodology, resting on a rigorous foundation, and going beyond the \emph{ad hoc} approaches commonly found in classical matrix analysis.} In the latter, derivatives are typically obtained through differential identities or perturbation arguments without a fully rigorous treatment of differentiability. 
For instance, \citet{Horn2012} present differential identities for spectral functions, but these do not constitute a unified framework for higher-order derivatives.
\end{remark}

\section{Special Case ($n=1$): Closed-Form Singular-Value Jacobian}
\label{sec:special_case_jacobian}

We now show that Theorem~\ref{thm:nth_order_singular_value}~(\nameref{thm:nth_order_singular_value}) can recover the well-known singular-value Jacobian, stated in Lemma~\ref{lemma:jacobian_of_singular_value} \citep{StewartSun1990,magnusmatrix}.

\begin{lemma}[Closed-Form Singular-Value Jacobian]
\label{lemma:jacobian_of_singular_value}
The Jacobian of a singular value is well‐known in the literature \citep{StewartSun1990,magnusmatrix} in the form:
\begin{align}
\frac{\partial \sigma_k}{\partial A}
= u_k \, v_k^\top
,
\end{align}
which immediately admits an equivalent result with Kronecker-product presentation:
\begin{align}
D\sigma_k[\dd A] =
(v_k \otimes u_k)^\top \mathrm{vec}(\dd A).
\end{align}
\end{lemma}

\bigskip
\noindent
\textbf{Traditional Method in Matrix Analysis.} In classical matrix analysis \citep{StewartSun1990,magnusmatrix}, the derivation of singular-value derivatives often begins with the identity
\begin{align}
\sigma_k = u_k^\top A v_k,
\end{align}
and then applies the trace identity
\begin{align}
\sigma_k = \tr(u_k^\top A v_k),
\end{align}
to compute $\dd \sigma_k$ and its derivatives. However, this approach is largely ad hoc and does not scale systematically to higher-order derivatives or more general operator settings.

\bigskip
\begin{mynewproof}

We follow the schematic procedure suggested by Remark~\ref{remark:schematic_procedure_singular_value_derivative}~(\nameref{remark:schematic_procedure_singular_value_derivative}) to recover this first-order singular-value Jacobian by specializing $n=1$ in Theorem~\ref{thm:nth_order_singular_value}~(\nameref{thm:nth_order_singular_value}).

\bigskip
\noindent
\textbf{Procedure I -- Construct Infinitesimal Perturbation.}
We construct a perturbation series on $A$:
\begin{align}
    A(x) = A + xA^{(1)}, \quad  A^{(0)}=A
    ,
\end{align}
so it yields:
\begin{align}
\dd A = \lim_{x \to 0 }xA^{(1)} 
.    
\end{align}

\bigskip
\noindent
\textbf{Procedure II -- Specialize $n$ to Obtain Derivative Operator.} Specializing $n=1$ in Theorem~\ref{thm:nth_order_singular_value}~(\nameref{thm:nth_order_singular_value}) yields:
\begin{align}
\sigma_k^{(1)} = \langle w_k^{(+0)}, \mathcal{T}^{(1)} w_k^{(0)} \rangle.
\end{align}

\bigskip
\noindent
\textbf{Simplifying First-Order Term.}
Consider:
\begin{align}
&\mathcal{T}^{(1)} w_k^{(+0)} = 
\begin{bmatrix} 
0 & A^{(1)} \\ (A^{(1)})^\top & 0 
\end{bmatrix} 
\cdot \frac{1}{\sqrt{2}} 
\begin{bmatrix} 
u_k \\ 
v_k \end{bmatrix} =
\frac{1}{\sqrt{2}} \begin{bmatrix} A^{(1)} v_k \\ (A^{(1)})^\top u_k \end{bmatrix} \\
&\xLongrightarrow[]{}
\langle w_k^{(+0)}, \mathcal{T}^{(1)} w_k^{(+0)} \rangle = 
\left( \frac{1}{\sqrt{2}} 
\begin{bmatrix} u_k \\ v_k \end{bmatrix} \right)^\top \cdot \frac{1}{\sqrt{2}} \begin{bmatrix} A^{(1)} v_k \\ (A^{(1)})^\top u_k \end{bmatrix} \\
&\xLongrightarrow[]{}
\langle w_k^{(+0)}, \mathcal{T}^{(1)} w_k^{(+0)} \rangle = 
\frac{1}{2} \left[ u_k^\top A^{(1)} v_k + v_k^\top (A^{(1)})^\top u_k \right] \\
&\xLongrightarrow[]{}
\langle w_k^{(+0)}, \mathcal{T}^{(1)} w_k^{(+0)} \rangle = u_k^\top A^{(1)} v_k \\
&\xLongrightarrow[]{}
\sigma_k^{(1)} = u_k^\top A^{(1)} v_k .
\end{align}

\bigskip
\noindent
\textbf{Procedure III -- Map Derivative Operator Layout.} By Theorem~\ref{thm:frechet_taylor_operator}~(\nameref{thm:frechet_taylor_operator}), we have:
\begin{align}
   D \sigma_k [\dd A] 
   &= \sigma_k^{(1)}[\dd A] \\
   &= \lim_{x \to 0} \sigma_k^{(1)} x A^{(1)} \\
   &= \lim_{x \to 0} u_k^T xA^{(1)} v_k \\
   &=u_k^{\top} \dd A v_k  \in \mathbb{R}.
\end{align}

Using following identities from Lemma~\ref{lemma:essential_matrix_identities}~(\nameref{lemma:essential_matrix_identities}):
\begin{enumerate}
    \item $\tr(x)=x$,
    
    \item $\operatorname{vec}(B V A^\top) = (A \otimes B)\operatorname{vec}(V)$,

    \item $(A \otimes B)^\top = A^\top \otimes B^\top$,
    
    \item $\tr(ABC) = \tr(CAB) = \tr(BCA)$,
\end{enumerate}
yields:
\begin{align}
D \sigma_k [\dd A] 
   &= \tr(u_k^{\top} \dd A v_k) \\
   &= \tr(v_k u_k^{\top} \dd A ) \\
   &= (v_k \otimes u_k)^\top \mathrm{vec}(\dd A)
   ,
\end{align}
and:
\begin{align}
\frac{\partial \sigma_k}{\partial A}
    =(D\sigma_k)^\top = u_kv_k^\top.
\end{align}

{\hfill$\qed$\par}
\end{mynewproof}

\section{Special Case ($n=2$): Closed-Form Singular-Value Hessian}
\label{sec:special_case_hessian}

Explicit closed-form expressions for the singular-value Hessian of rectangular matrices are, to the best of our knowledge, not available in the literature. Such a result is essential for applications in stochastic analysis, for example when applying It\^o's lemma to stochastic differential equations (SDEs) or stochastic partial differential equations (SPDEs) driven by Wiener processes \citep{oksendal2003stochastic}. We now derive the singular-value Hessian for general real rectangular matrices, under Assumption~\ref{assump:simple_singular_values}~(\nameref{assump:simple_singular_values}), as stated in Lemma~\ref{lemma:hessian_of_singular_value}~(\nameref{lemma:hessian_of_singular_value}), represented in the layout:
\begin{align}
\operatorname{vec}(\dd A)^\top
\left(
\frac{\partial}{\partial \operatorname{vec}(A)} 
\operatorname{vec}\!\left( \frac{\partial \sigma_k}{\partial A}\right)
\right)
\operatorname{vec}(\dd A),
\end{align}
by specializing Theorem~\ref{thm:nth_order_singular_value}~(\nameref{thm:nth_order_singular_value}) to the case $n=2$.

\begin{lemma}[Closed-Form Singular-Value Hessian]
\label{lemma:hessian_of_singular_value}
%Define scale factor and its inverse
The Hessian of a singular value is given as:\\
\pgfmathsetmacro{\linewidthfactor}{0.9}
\pgfmathsetmacro{\inverselinewidthfactor}{1/\linewidthfactor}
\scalebox{\linewidthfactor}{% scale
\begin{minipage}{\inverselinewidthfactor\linewidth} % 1/scale
\begin{align}
\frac{\partial}{\partial \operatorname{vec}(A)} \operatorname{vec}\left(
   \frac{\partial \sigma_k}{\partial A}
   \right) 
&=\underbrace{\sum_{i \neq k, i \leq m} \frac{\sigma_k}{\sigma_k^2 - \sigma_i^2}\left(v_k \otimes u_i\right)\left(v_k \otimes u_i\right)^\top}_{\text{left}} +\\
    &\qquad \underbrace{\sum_{j \neq k, j \leq n} \frac{\sigma_k}{\sigma_k^2 - \sigma_j^2} \left(v_j \otimes u_k\right)\left(v_j \otimes u_k\right)^\top}_{\text{right}} + \\
    &\qquad \underbrace{\sum_{l \neq k, l \leq r} \frac{\sigma_l}{\sigma_k^2 - \sigma_l^2}\left[ \left(v_k \otimes u_l\right)\left(v_l \otimes u_k\right)^\top + \left(v_l \otimes u_k\right)\left(v_k \otimes u_l\right)^\top \right]}_{\text{left-right interaction}} 
\end{align}
\end{minipage}%
}\\
with Kronecker-product representation.
\end{lemma}

\begin{mynewproof}

We follow the schematic procedure suggested by Remark~\ref{remark:schematic_procedure_singular_value_derivative}~(\nameref{remark:schematic_procedure_singular_value_derivative}) to derive this second-order singular-value Hessian by specializing $n=2$ in Theorem~\ref{thm:nth_order_singular_value}~(\nameref{thm:nth_order_singular_value}).

\bigskip
\noindent
\textbf{Procedure I -- Construct Infinitesimal Perturbation.} We construct a perturbation series on $A$:
\begin{align}
    A(x) = A + xA^{(1)}, \quad  A^{(0)}=A
    ,
\end{align}
so it yields:
\begin{align}
\dd A = \lim_{x \to 0 }xA^{(1)} 
.    
\end{align}

\bigskip
\noindent
\textbf{Procedure II -- Specialize $n$ to Obtain Derivative Operator.} Specializing $n=2$ in Theorem~\ref{thm:nth_order_singular_value}~(\nameref{thm:nth_order_singular_value}) yields:
\begin{align}
\sigma_k^{(2)} &= \sum_{p=1}^2 (-1)^{p-1} \sum_{\substack{i_1 + \cdots + i_p = 2 \\ i_j \geq 1}} \langle w_k^{(+0)}, \mathcal{T}^{(i_1)} S_k^{(0)} \mathcal{T}^{(i_2)} \cdots S_k^{(0)} \mathcal{T}^{(i_p)} w_k^{(+0)} \rangle  \\
&=\sigma_k^{(2,p=1)} + \sigma_k^{(2,p=2)}
\end{align}
where
\begin{align}
    \sigma_k^{(2,p=1)} := \langle w_k^{(+0)}, \mathcal{T}^{(2)} w_k^{(+0)} \rangle,
\end{align}
and
\begin{align}
\sigma_k^{(2, p=2)} := - \langle w_k^{(+0)}, \mathcal{T}^{(1)} S_k^{(0)} \mathcal{T}^{(1)} w_k^{(+0)} \rangle .
\end{align}

\bigskip
\noindent
\textbf{Computing Term $\sigma_k^{(2,p=1)}$.} We first compute the term with $p=1$ ($\sigma_k^{(2,p=1)}$). By Theorem~\ref{thm:nth_order_singular_value}~(\nameref{thm:nth_order_singular_value}), we substitute:
\begin{align}
\mathcal{T}^{(2)}=    
\begin{bmatrix} 
0 & A^{(2)} \\ 
(A^{(2)})^\top & 0 
\end{bmatrix}
\quad
\text{and} \quad
w_k^{(+0)} = \frac{1}{\sqrt{2}}
\begin{bmatrix} 
u_k \\ 
v_k 
\end{bmatrix}
\end{align}
into:
\begin{align}
\mathcal{T}^{(2)} w_k^{(+0)},
\end{align}
it yields:
\begin{align}
&\mathcal{T}^{(2)} w_k^{(+0)} = 
\begin{bmatrix} 
0 & A^{(2)} \\ 
(A^{(2)})^\top & 0 
\end{bmatrix} 
\cdot \frac{1}{\sqrt{2}} 
\begin{bmatrix} 
u_k \\ 
v_k 
\end{bmatrix} = \frac{1}{\sqrt{2}} 
\begin{bmatrix} 
A^{(2)} v_k \\ 
(A^{(2)})^\top u_k 
\end{bmatrix} 
.
\end{align}

Substituting 
\begin{align}
\mathcal{T}^{(2)} w_k^{(+0)} = \frac{1}{\sqrt{2}} 
\begin{bmatrix} 
A^{(2)} v_k \\ 
(A^{(2)})^\top u_k 
\end{bmatrix} 
\end{align}
into:
\begin{align}
\sigma_k^{(2,p=1)} = \langle w_k^{(+0)}, \mathcal{T}^{(2)} w_k^{(+0)} \rangle
\end{align}
yields:
\begin{align}
\sigma_k^{(2,p=1)} &= \langle w_k^{(+0)}, \mathcal{T}^{(2)} w_k^{(+0)} \rangle \\
&= \frac{1}{2} \left[ u_k^\top A^{(2)} v_k + (v_k)^\top (A^{(2)})^\top u_k \right] \\
&= u_k^\top A^{(2)} v_k 
\end{align}

In the construction of $\dd A$, there is:
\begin{align}
    A^{(2)} = O,
\end{align}
so that:
\begin{align}
\sigma_k^{(2, p=1)} = u_k^\top A^{(2)} v_k = 0.   
\end{align}

\bigskip
\noindent
\textbf{Sketch of Computing Term $\sigma_k^{(2,p=2)}$.} We compute the term with $p=2$:
\begin{align}
\sigma_k^{(2, p=2)} = - \langle w_k^{(+0)}, \mathcal{T}^{(1)} S_k^{(0)} \mathcal{T}^{(1)} w_k^{(+0)} \rangle \label{equ:hessian_term_p_2}
.
\end{align}
To simplify, we first compute:
\begin{align}
S_k^{(0)} \mathcal{T}^{(1)} w_k^{(+0)} \label{equ:hessian_STw}
,
\end{align}
then substitute this result into Equation~(\ref{equ:hessian_term_p_2}) to produce complete $\sigma_k^{(2, p=2)}$.

\bigskip
\noindent
\textbf{Computing Contributions in $S_k^{(0)} \mathcal{T}^{(1)} w_k^{(+0)} $ in $\sigma_k^{(2,p=2)}$.}~By Theorem~\ref{thm:nth_order_singular_value}~(\nameref{thm:nth_order_singular_value}), the unperturbed reduced resolvent is defined as:%\\
% \pgfmathsetmacro{\linewidthfactor}{0.9}
% \pgfmathsetmacro{\inverselinewidthfactor}{1/\linewidthfactor}
% \scalebox{\linewidthfactor}{% scale
% \begin{minipage}{\inverselinewidthfactor\linewidth} % 1/scale
\begin{align}
S_k^{(0)} &= 
\sum_{i=1,i \neq k}^{r} \frac{w_i^{(+0)} (w_i^{(+0)})^\top}{\sigma_i^{(0)} - \sigma_k^{(0)}} + 
\sum_{i=1,i \neq k}^{r} \frac{w_i^{(-0)} (w_i^{(-0)})^\top}{-\sigma_i^{(0)} - \sigma_k^{(0)}} - 
\sum_{j=r+1}^m \frac{a_j^{(0)} (a_j^{(0)})^\top}{\sigma_k^{(0)}} 
- 
\sum_{j=r+1}^n \frac{b_j^{(0)} (b_j^{(0)})^\top}{\sigma_k^{(0)}} 
.
\end{align}
% \end{minipage}%
% }

By Theorem~\ref{thm:nth_order_singular_value}~(\nameref{thm:nth_order_singular_value}), substituting non-null eigenvectors of the unperturbed embedding $\mathcal{T}^{(0)}$:
\begin{align}
w_i^{(+0)} = \frac{1}{\sqrt{2}} 
\begin{bmatrix} 
u_i \\ 
v_i 
\end{bmatrix}
, 
\quad w_i^{(-0)} = \frac{1}{\sqrt{2}} 
\begin{bmatrix} 
u_i \\ 
-v_i 
\end{bmatrix}
,    
\end{align}
and null eigenvectors of the unperturbed embedding $\mathcal{T}^{(0)}$: 
\begin{align}
a_j^{(0)} = 
\begin{bmatrix} 
u_j \\ 
0 
\end{bmatrix}  
, \quad
b_j^{(0)}=
\begin{bmatrix} 
0 
\\ 
v_j 
\end{bmatrix}
, 
\end{align}
into $S_k^{(0)}$ yields:\\
\pgfmathsetmacro{\linewidthfactor}{0.9}
\pgfmathsetmacro{\inverselinewidthfactor}{1/\linewidthfactor}
\scalebox{\linewidthfactor}{% scale
\begin{minipage}{\inverselinewidthfactor\linewidth} % 1/scale
\begin{align}
S_k^{(0)} =
\underbrace{
\sum_{i=1, i \neq k}^{r} \frac{1}{2}\frac{
\begin{bmatrix}
    u_i \\
    v_i
\end{bmatrix}
\begin{bmatrix}
u_i^\top \; v_i^\top
\end{bmatrix}
}{\sigma_i - \sigma_k}
}_{\text{positive eigenspaces}}
+
\underbrace{
\sum_{i=1, i \neq k}^{r} \frac{1}{2}\frac{
\begin{bmatrix}
    u_i \\
    -v_i
\end{bmatrix}
\begin{bmatrix}
u_i^\top \; -v_i^\top
\end{bmatrix}
}{-\sigma_i - \sigma_k}
}_{\text{negative eigenspaces}}
-
\underbrace{
\sum_{j=r+1}^m \frac{
\begin{bmatrix}
    u_j \\
    0
\end{bmatrix}
\begin{bmatrix}
u_j^\top \; 0
\end{bmatrix}
}{\sigma_k^{(0)}} 
}_{\text{left-null eigenspaces}}
-
\underbrace{
\sum_{j=r+1}^n \frac{
\begin{bmatrix}
    0 \\
    v_j
\end{bmatrix}
\begin{bmatrix}
0 \; v_j^\top
\end{bmatrix}
}{\sigma_k^{(0)}}
}_{\text{right-null eigenspaces}}
,
\end{align}
\end{minipage}
}\\
where:
\begin{enumerate}
    \item \textbf{contributions in positive eigenspaces} ($S_k^{(+0)}$) represents the contribution in the subspaces associated with $w_{i}^{(+0)}$;

    \item \textbf{contributions in negative eigenspaces} ($S_k^{(-0)}$) represents the contribution in the subspaces associated with $w_{i}^{(-0)}$;

    \item \textbf{contributions in left-null eigenspaces} ($S_k^{(0,a)}$) represents the contribution in the subspaces associated with $a_j^{(0)}$;

    \item \textbf{contributions in right-null eigenspaces} ($S_k^{(0,b)}$) represents the contribution in the subspaces associated with $b_j^{(0)}$.
\end{enumerate}

Substituting $w_k^{(+0)}$ into:
\begin{align}
S_k^{(0)}\mathcal{T}^{(1)} w_k^{(+0)}   
\end{align}
yields:
\begin{align}
S_k^{(0)}\mathcal{T}^{(1)} w_k^{(+0)} = S_k^{(0)}\mathcal{T}^{(1)}
\frac{1}{\sqrt{2}} 
\begin{bmatrix} 
u_i \\ 
v_i 
\end{bmatrix} 
= S_k^{(0)}\frac{1}{\sqrt{2}} 
\begin{bmatrix} 
A^{(1)} v_k \\ 
(A^{(1)})^\top u_k
\end{bmatrix} 
,
\end{align}
then apply the explicit $S_k^{(0)}$ on this result:
\begin{align}
S_k^{(0)}\mathcal{T}^{(1)} w_k^{(+0)} = S_k^{(0)}\frac{1}{\sqrt{2}} 
\begin{bmatrix} 
A^{(1)} v_k \\ 
(A^{(1)})^\top u_k
\end{bmatrix} 
=(S_k^{(+0)} + S_k^{(-0)} + S_k^{(0,a)}+S_k^{(0,b)})\frac{1}{\sqrt{2}} 
\begin{bmatrix} 
A^{(1)} v_k \\ 
(A^{(1)})^\top u_k
\end{bmatrix} 
,
\end{align}
and discuss the contributions in terms of subspaces:
\begin{enumerate}[leftmargin=*]

\item \textbf{Contributions in Positive Eigenspaces}. 
To compute
\begin{align}
S_k^{(+0)}
\frac{1}{\sqrt{2}} 
\begin{bmatrix} 
A^{(1)} v_k \\ 
(A^{(1)})^\top u_k
\end{bmatrix}
&= \sum_{i=1,i\neq k}^r 
\frac{w_i^{(+0)} (w_i^{(+0)})^\top}{\sigma_i - \sigma_k} 
\frac{1}{\sqrt{2}} 
\begin{bmatrix} 
A^{(1)} v_k \\ 
(A^{(1)})^\top u_k
\end{bmatrix} 
,
\end{align}
consider:
\begin{align}
(w_i^{(+0)})^\top \cdot \frac{1}{\sqrt{2}} 
\begin{bmatrix} 
A^{(1)} v_k \\ 
(A^{(1)})^\top u_k 
\end{bmatrix} &= 
\frac{1}{\sqrt{2}} 
\begin{bmatrix} 
u_i^\top & v_i^\top 
\end{bmatrix} 
\cdot \frac{1}{\sqrt{2}} 
\begin{bmatrix} 
A^{(1)} v_k \\ 
(A^{(1)})^\top u_k 
\end{bmatrix} \notag \\
&= \frac{1}{2} \left[ u_i^\top A^{(1)} v_k + v_i^\top (A^{(1)})^\top u_k \right] .
\end{align}

Note the identity:
\begin{align}
v_i^\top (A^{(1)})^\top u_k = u_k^\top A^{(1)} v_i    \in \mathbb{R}
,
\end{align}
so that:
\begin{align}
&
(w_i^{(+)})^\top \cdot \frac{1}{\sqrt{2}} 
\begin{bmatrix} 
A^{(1)} v_k \\ 
(A^{(1)})^\top u_k 
\end{bmatrix} 
=
\frac{1}{2} \left[ u_i^\top A^{(1)} v_k + u_k^\top A^{(1)} v_i \right].
\end{align}

Therefore the contributions in positive eigenspaces are given as:
\begin{align}
S_k^{(+0)}
\frac{1}{\sqrt{2}} 
\begin{bmatrix} 
A^{(1)} v_k \\ 
(A^{(1)})^\top u_k
\end{bmatrix}
&= \sum_{i=1,i\neq k}^r 
\frac{w_i^{(+0)} (w_i^{(+0)})^\top}{\sigma_i - \sigma_k} 
\frac{1}{\sqrt{2}} 
\begin{bmatrix} 
A^{(1)} v_k \\ 
(A^{(1)})^\top u_k
\end{bmatrix} \\
&= \sum_{i=1,i\neq k}^r  
\frac{1}{\sqrt{2}} \cdot \frac{u_i^\top A^{(1)} v_k + u_k^\top A^{(1)} v_i}{2 (\sigma_i - \sigma_k)} \begin{bmatrix} u_i \\ v_i \end{bmatrix}.    
\end{align}

\item \textbf{Contributions in Negative Eigenspaces}.

To compute:
\begin{align}
S_k^{(-0)}
\frac{1}{\sqrt{2}} 
\begin{bmatrix} 
A^{(1)} v_k \\ 
(A^{(1)})^\top u_k
\end{bmatrix}
&= \sum_{i=1,i\neq k}^r 
\frac{w_i^{(-0)} (w_i^{(-0)})^\top}{-\sigma_i - \sigma_k} 
\frac{1}{\sqrt{2}} 
\begin{bmatrix} 
A^{(1)} v_k \\ 
(A^{(1)})^\top u_k
\end{bmatrix} 
,
\end{align}
consider:
\begin{align}
(w_i^{(-0)})^\top \cdot \frac{1}{\sqrt{2}} 
\begin{bmatrix} 
A^{(1)} v_k \\ 
(A^{(1)})^\top u_k 
\end{bmatrix} 
&= 
\frac{1}{\sqrt{2}} 
\begin{bmatrix} 
u_i^\top & -v_i^\top 
\end{bmatrix} 
\cdot \frac{1}{\sqrt{2}} 
\begin{bmatrix} 
A^{(1)} v_k \\ 
(A^{(1)})^\top u_k 
\end{bmatrix} \notag \\
&=\frac{1}{2} \left[ u_i^\top A^{(1)} v_k - v_i^\top (A^{(1)})^\top u_k \right].
\end{align}

Note the identity:
\begin{align}
v_i^\top (A^{(1)})^\top u_k = u_k^\top A^{(1)} v_i    \in \mathbb{R}
\end{align}
so that:
\begin{align}
(w_i^{(+0)})^\top \cdot \frac{1}{\sqrt{2}} 
\begin{bmatrix} 
A^{(1)} v_k \\ 
(A^{(1)})^\top u_k 
\end{bmatrix} 
&=
\frac{1}{2} \left[ u_i^\top A^{(1)} v_k - u_k^\top A^{(1)} v_i \right] \\
&= 
\frac{1}{\sqrt{2}} \cdot \frac{u_i^\top A^{(1)} v_k - u_k^\top A^{(1)} v_i}{2 (-\sigma_i - \sigma_k)} 
\begin{bmatrix} 
u_i \\ 
-v_i \end{bmatrix}.   
\end{align}

Therefore the contributions in negative eigenspaces are given as:
\begin{align}
S_k^{(-0)}
\frac{1}{\sqrt{2}} 
\begin{bmatrix} 
A^{(1)} v_k \\ 
(A^{(1)})^\top u_k
\end{bmatrix}
&= \sum_{i=1,i\neq k}^r 
\frac{w_i^{(-0)} (w_i^{(-0)})^\top}{-\sigma_i - \sigma_k} 
\frac{1}{\sqrt{2}} 
\begin{bmatrix} 
A^{(1)} v_k \\ 
(A^{(1)})^\top u_k
\end{bmatrix} \\
&=\sum_{i=1,i\neq k}^r 
\frac{1}{\sqrt{2}} \cdot \frac{u_i^\top A^{(1)} v_k - u_k^\top A^{(1)} v_i}{2 (-\sigma_i - \sigma_k)} 
\begin{bmatrix} 
u_i \\ 
-v_i \end{bmatrix}.
\end{align}

\item \textbf{Contributions in Left-Null Eigenspaces}.

To compute:
\begin{align}
S_k^{(0,a)}
\frac{1}{\sqrt{2}} 
\begin{bmatrix} 
A^{(1)} v_k \\ 
(A^{(1)})^\top u_k
\end{bmatrix}
&= -\sum_{j=r+1}^{m} 
\frac{a_j^{(0)} (a_j^{(0)})^\top}{ \sigma_k} 
\frac{1}{\sqrt{2}} 
\begin{bmatrix} 
A^{(1)} v_k \\ 
(A^{(1)})^\top u_k
\end{bmatrix} 
,
\end{align}
consider:
\begin{align}
(a_j^{(0)})^\top \cdot \frac{1}{\sqrt{2}} 
\begin{bmatrix} 
A^{(1)} v_k \\ 
(A^{(1)})^\top u_k 
\end{bmatrix} 
&= 
\frac{1}{\sqrt{2}} 
\begin{bmatrix} 
u_j^\top & 0 
\end{bmatrix} 
\cdot \frac{1}{\sqrt{2}} 
\begin{bmatrix} 
A^{(1)} v_k \\ 
(A^{(1)})^\top u_k 
\end{bmatrix} \notag \\
&=\frac{1}{2} u_j^\top A^{(1)} v_k .
\end{align}

Therefore the contributions in left-null eigenspaces are given as:
\begin{align}
S_k^{(0,a)}
\frac{1}{\sqrt{2}} 
\begin{bmatrix} 
A^{(1)} v_k \\ 
(A^{(1)})^\top u_k
\end{bmatrix}
&= -\sum_{j=r+1}^{m} 
\frac{a_j^{(0)} (a_j^{(0)})^\top}{ \sigma_k} 
\frac{1}{\sqrt{2}} 
\begin{bmatrix} 
A^{(1)} v_k \\ 
(A^{(1)})^\top u_k
\end{bmatrix} \\
&=-\sum_{j=r+1}^{m} 
\frac{1}{\sqrt{2}} \cdot \frac{u_j^\top A^{(1)} v_k}{\sigma_k} \begin{bmatrix} 
u_j \\ 
0 
\end{bmatrix}.
\end{align}

\item \textbf{Contributions in Right-Null Eigenspaces}.

To compute:
\begin{align}
S_k^{(0,b)}
\frac{1}{\sqrt{2}} 
\begin{bmatrix} 
A^{(1)} v_k \\ 
(A^{(1)})^\top u_k
\end{bmatrix}
&= -\sum_{j=r+1}^{n} 
\frac{b_j^{(0)} (b_j^{(0)})^\top}{ \sigma_k} 
\frac{1}{\sqrt{2}} 
\begin{bmatrix} 
A^{(1)} v_k \\ 
(A^{(1)})^\top u_k
\end{bmatrix} 
,
\end{align}
consider:
\begin{align}
(b_j^{(0)})^\top \cdot \frac{1}{\sqrt{2}} 
\begin{bmatrix} 
A^{(1)} v_k \\ 
(A^{(1)})^\top u_k 
\end{bmatrix} 
&= 
\frac{1}{\sqrt{2}} 
\begin{bmatrix} 
0 & v_j^\top 
\end{bmatrix} 
\cdot \frac{1}{\sqrt{2}} 
\begin{bmatrix} 
A^{(1)} v_k \\ 
(A^{(1)})^\top u_k 
\end{bmatrix} \notag \\
&=\frac{1}{2} v_j^\top (A^{(1)})^\top u_k \\
&= \frac{1}{2} u_k^\top A^{(1)} v_j.
\end{align}

Therefore the contributions in right-null eigenspaces are given as:
\begin{align}
 S_k^{(0,b)}
\frac{1}{\sqrt{2}} 
\begin{bmatrix} 
A^{(1)} v_k \\ 
(A^{(1)})^\top u_k
\end{bmatrix}
&= -\sum_{j=r+1}^{n} 
\frac{b_j^{(0)} (b_j^{(0)})^\top}{ \sigma_k} 
\frac{1}{\sqrt{2}} 
\begin{bmatrix} 
A^{(1)} v_k \\ 
(A^{(1)})^\top u_k
\end{bmatrix} \\
&=-\sum_{j=r+1}^{n} 
\frac{1}{\sqrt{2}} \cdot \frac{u_k^\top A^{(1)} v_j}{\sigma_k} \begin{bmatrix} 
0 \\ 
v_j 
\end{bmatrix}.
\end{align}

\end{enumerate}

\bigskip
\noindent
\textbf{Computing Inner-Product Contributions in $\langle w_k, \mathcal{T}^{(1)} S_k \mathcal{T}^{(1)} w_k \rangle$.} Since the term:
\begin{align}
 S_k^{(0)} \mathcal{T}^{(1)} w_k^{(+0)}    
\end{align}
is computed above, to further derive:
\begin{align}
 \sigma_k^{(2, p=2)} = - \langle w_k^{(+0)}, \mathcal{T}^{(1)} S_k^{(0)} \mathcal{T}^{(1)} w_k^{(+0)} \rangle
 ,
\end{align}
we compute the inner-product contributions in $\langle w_k, \mathcal{T}^{(1)} S_k \mathcal{T}^{(1)} w_k \rangle$ with respect to subspaces:
\begin{enumerate}[leftmargin=*]

\item \textbf{Inner-Product Contributions in Positive Eigenspaces}. 
% \pgfmathsetmacro{\linewidthfactor}{0.75}
% \pgfmathsetmacro{\inverselinewidthfactor}{1/\linewidthfactor}
% \scalebox{\linewidthfactor}{% scale
% \begin{minipage}{\inverselinewidthfactor\linewidth} % 1/scale
Consider:
\begin{align}
    Z^{(+0)} &:= \mathcal{T}^{(1)} S_k^{(+0)} \mathcal{T}^{(1)} w_k^{(+0)} \\
    &= \mathcal{T}^{(1)}  
    \sum_{i=1,i\neq k}^r 
\frac{1}{\sqrt{2}} \cdot \frac{u_i^\top A^{(1)} v_k + u_k^\top A^{(1)} v_i}{2 (\sigma_i - \sigma_k)} \begin{bmatrix} u_i \\ v_i \end{bmatrix} \\
&=\sum_{i=1,i\neq k}^r \frac{1}{\sqrt{2}} \cdot \frac{u_i^\top A^{(1)} v_k + u_k^\top A^{(1)} v_i}{2 (\sigma_i - \sigma_k)} \begin{bmatrix} 
A^{(1)} v_i \\ 
(A^{(1)})^\top u_i 
\end{bmatrix}
,
\end{align}
so that:\\
\pgfmathsetmacro{\linewidthfactor}{0.9}
\pgfmathsetmacro{\inverselinewidthfactor}{1/\linewidthfactor}
\scalebox{\linewidthfactor}{% scale
\begin{minipage}{\inverselinewidthfactor\linewidth} % 1/scale
\begin{align}
\langle 
w_k, Z^{(+0)}
\rangle
&=
\sum_{i\neq k}
\left(
\frac{1}{\sqrt{2}}
\begin{bmatrix}
    u_k \\
    v_k
\end{bmatrix}
\right)^\top
\frac{1}{\sqrt{2}} \cdot \frac{u_i^\top A^{(1)} v_k + u_k^\top A^{(1)} v_i}{2 (\sigma_i - \sigma_k)} \begin{bmatrix} 
A^{(1)} v_i \\ 
(A^{(1)})^\top u_i 
\end{bmatrix} \\
&=\sum_{i=1,i\neq k}^r
\left(\frac{1}{\sqrt{2}}\right)^2 \cdot \frac{u_i^\top A^{(1)} v_k + u_k^\top A^{(1)} v_i}{2 (\sigma_i - \sigma_k)} \cdot  \left[ u_k^\top A^{(1)} v_i + v_k^\top (A^{(1)})^\top u_i \right] \\
&= \sum_{i=1,i\neq k}^r
\frac{1}{4 (\sigma_i - \sigma_k)} \left[ u_i^\top A^{(1)} v_k + u_k^\top A^{(1)} v_i \right] \left[ u_k^\top A^{(1)} v_i + u_i^\top A^{(1)} v_k \right] \\
&= \sum_{i=1,i\neq k}^r
\frac{1}{4 (\sigma_i - \sigma_k)} \left[ u_i^\top A^{(1)} v_k + u_k^\top A^{(1)} v_i \right]^2  \\
&=
\sum_{i=1,i\neq k}^r
\frac{1}{4 (\sigma_i - \sigma_k)} 
\left[ \left[ u_i^\top A^{(1)} v_k \right]^2 + 2 u_i^\top A^{(1)} v_k u_k^\top A^{(1)} v_i + \left[ u_k^\top A^{(1)} v_i \right]^2 \right] 
.
\end{align}
\end{minipage}
}\\

\item \textbf{Inner-Product Contributions in Negative Eigenspaces}. Consider:
\begin{align}
    Z^{(-0)} &:= \mathcal{T}^{(1)} S_k^{(-0)} \mathcal{T}^{(1)} w_k^{(+0)} \\
    &=\mathcal{T}^{(1)}  \sum_{i=1,i\neq k}^r  \frac{1}{\sqrt{2}} \cdot \frac{u_i^\top A^{(1)} v_k - u_k^\top A^{(1)} v_i}{2 (-\sigma_i - \sigma_k)} 
\begin{bmatrix} 
u_i \\ 
-v_i 
\end{bmatrix} \\
&=\sum_{i=1,i\neq k}^r
\frac{1}{\sqrt{2}} \cdot \frac{u_i^\top A^{(1)} v_k - u_k^\top A^{(1)} v_i}{2 (-\sigma_i - \sigma_k)} \begin{bmatrix} A^{(1)} (-v_i) \\ (A^{(1)})^\top u_i \end{bmatrix}
,
\end{align}
so that:\\
\pgfmathsetmacro{\linewidthfactor}{0.9}
\pgfmathsetmacro{\inverselinewidthfactor}{1/\linewidthfactor}
\scalebox{\linewidthfactor}{% scale
\begin{minipage}{\inverselinewidthfactor\linewidth} % 1/scale
\begin{align}
\langle w_k, Z^{(-0)} \rangle &= \sum_{i=1,i\neq k}^r \frac{1}{2} \cdot \frac{u_i^\top A^{(1)} v_k - u_k^\top A^{(1)} v_i}{2 (-\sigma_i - \sigma_k)} \left[ u_k^\top A^{(1)} (-v_i) + v_k^\top (A^{(1)})^\top u_i \right] \\
&=\sum_{i=1,i\neq k}^r 
\frac{1}{4 (-\sigma_i - \sigma_k)} \left[ u_i^\top A^{(1)} v_k - u_k^\top A^{(1)} v_i \right] \left[ -(u_k)^\top A^{(1)} v_i + u_i^\top A^{(1)} v_k \right] \\
&=\sum_{i=1,i\neq k}^r
\frac{1}{4 (-\sigma_i - \sigma_k)} 
\left[ u_i^\top A^{(1)} v_k - u_k^\top A^{(1)} v_i \right]^2 \\
&=\sum_{i=1,i\neq k}^r
\frac{1}{4 (-\sigma_i - \sigma_k)} 
\left[ \left[ u_i^\top A^{(1)} v_k \right]^2 - 2 u_i^\top A^{(1)} v_k u_k^\top A^{(1)} v_i + \left[ u_k^\top A^{(1)} v_i \right]^2 \right] %\\
% &\xLongrightarrow[]{\text{sum over}~i \neq k}
% \sum_{i=1}^r \frac{1}{4 (-\sigma_i -\sigma_k)} 
% \left[ \left[ u_i^\top A^{(1)} v_k \right]^2 - 2 u_i^\top A^{(1)} v_k u_k^\top A^{(1)} v_i + \left[ u_k^\top A^{(1)} v_i \right]^2 \right]
.
\end{align}
\end{minipage}
}

\item \textbf{Inner-Product Contributions in Left-Null Eigenspaces}. Consider:
\begin{align}
    Z^{(0,a)} &:= \mathcal{T}^{(1)} S_k^{(0,a)} \mathcal{T}^{(1)} w_k^{(+0)} \\
    &= \mathcal{T}^{(1)} \sum_{j=r+1}^m
    \left( -\frac{1}{\sqrt{2}} \cdot \frac{u_j^\top A^{(1)} v_k}{\sigma_k} \begin{bmatrix} 
u_j \\ 
0 
\end{bmatrix} \right) \\
&= \sum_{j=r+1}^m 
-\frac{1}{\sqrt{2}} \cdot \frac{u_j^\top A^{(1)} v_k}{\sigma_k} \begin{bmatrix} 
0 \\ 
(A^{(1)})^\top u_j 
\end{bmatrix}
,
\end{align}
so that:
\begin{align}
\langle w_k, Z^{(0,a)} \rangle 
&= \sum_{j=r+1}^m -\frac{1}{2} \cdot \frac{u_j^\top A^{(1)} v_k}{\sigma_k} \left[ u_k^\top \cdot 0 + v_k^\top (A^{(1)})^\top u_j \right]  \\
&=\sum_{j=r+1}^m
-\frac{1}{2} \cdot \frac{u_j^\top A^{(1)} v_k u_j^\top A^{(1)} v_k}{\sigma_k} 
.    
\end{align}

\item \textbf{Inner-Product Contributions in Right-Null Eigenspaces}. Consider:
\begin{align}
    Z^{(0,b)} &:= \mathcal{T}^{(1)} S_k^{(0,b)} \mathcal{T}^{(1)} w_k^{(+0)} \\
    &= \mathcal{T}^{(1)} 
    \sum_{j=r+1}^n
    \left( -\frac{1}{\sqrt{2}} \cdot \frac{u_k^\top A^{(1)} v_j}{\sigma_k} \begin{bmatrix} 0 \\ v_j \end{bmatrix} \right) \\
    &= \sum_{j=r+1}^n -\frac{1}{\sqrt{2}} \cdot \frac{u_k^\top A^{(1)} v_j}{\sigma_k} \begin{bmatrix} A^{(1)} v_j \\ 0 \end{bmatrix} 
    ,
\end{align}
so that:
\begin{align}
\langle w_k, Z^{(0,b)} \rangle &= 
\sum_{j=r+1}^n-\frac{1}{2} \cdot \frac{u_k^\top A^{(1)} v_j}{\sigma_k} \left[ u_k^\top A^{(1)} v_j + v_k^\top \cdot 0 \right]\\
&=\sum_{j=r+1}^n 
-\frac{1}{2} \cdot \frac{\left[ u_k^\top A^{(1)} v_j \right]^2}{\sigma_k} 
.    
\end{align}

\end{enumerate}

\bigskip
\noindent
\textbf{Combining Terms.} 
Since
\begin{align}
\sigma_k^{(2, p=1)} = 0,    
\end{align}
thus,
\begin{align}
\sigma_k^{(2)} &=\sigma_k^{(2, p=2)}\\
&= 
-\left[ 
\sum_{i=1,i \neq k}^r \frac{1}{4 (\sigma_i - \sigma_k)} 
\left[ \left[ u_i^\top A^{(1)} v_k \right]^2 + 2 u_i^\top A^{(1)} v_k u_k^\top A^{(1)} v_i + \left[ u_k^\top A^{(1)} v_i \right]^2 \right] \right. \\
&\qquad\left.+
\sum_{i=1,i \neq k}^r \frac{1}{4 (-\sigma_i - \sigma_k)} 
\left[ \left[ u_i^\top A^{(1)} v_k \right]^2 - 2 u_i^\top A^{(1)} v_k u_k^\top A^{(1)} v_i + \left[ u_k^\top A^{(1)} v_i \right]^2 \right] \right.\\
&\qquad\left.-\sum_{j=r+1}^m \frac{1}{2 \sigma_k} \left[ u_j^\top A^{(1)} v_k \right]^2
-\sum_{j=r+1}^n \frac{1}{2 \sigma_k} \left[ u_k^\top A^{(1)} v_j \right]^2
\right] \\
&= 
-\left[ 
\sum_{i=1,i \neq k}^r \frac{(\sigma_i + \sigma_k)  -(\sigma_i - \sigma_k)}{4 (\sigma_i^2 - \sigma_k^2)} 
\left[ \left[ u_i^\top A^{(1)} v_k \right]^2  + \left[ u_k^\top A^{(1)} v_i \right]^2 \right] \right. \\
&\qquad\left.
+\frac{1}{2}\sum_{i=1,i \neq k}^r \frac{(\sigma_i + \sigma_k) + (\sigma_i - \sigma_k)}{\sigma_i^2 - \sigma_k^2} u_i^\top A^{(1)} v_k u_k^\top A^{(1)} v_i
\right.\\
&\qquad\left.
-\sum_{j=r+1}^m \frac{1}{2 \sigma_k} \left[ u_j^\top A^{(1)} v_k \right]^2
-\sum_{j=r+1}^n \frac{1}{2 \sigma_k} \left[ u_k^\top A^{(1)} v_j \right]^2
\right] \\
&=\frac{1}{2} \sum_{i=1,i \neq k}^r \frac{\sigma_k}{\sigma_k^2 - \sigma_i^2} \left[ u_i^\top A^{(1)} v_k \right]^2 +
\frac{1}{2} \sum_{i \neq k}^r \frac{\sigma_k}{\sigma_k^2 - \sigma_i^2} \left[ u_k^\top A^{(1)} v_i \right]^2  \\
&\qquad+\frac{1}{2}\sum_{i=1,i \neq k}^r \frac{\sigma_i}{\sigma_k^2 - \sigma_i^2} u_i^\top A^{(1)} v_k u_k^\top A^{(1)} v_i +\frac{1}{2}\sum_{i=1}^r \frac{\sigma_i}{\sigma_k^2 - \sigma_i^2} u_i^\top A^{(1)} v_k u_k^\top A^{(1)} v_i\\
&\qquad 
+\frac{1}{2}\sum_{j=r+1}^m \frac{\sigma_k}{ \sigma_k^2-\sigma_j^2} \left[ u_j^\top A^{(1)} v_k \right]^2
+\frac{1}{2}\sum_{j=r+1}^n \frac{\sigma_k}{ \sigma_k^2-\sigma_j^2} \left[ u_k^\top A^{(1)} v_j \right]^2\\
&\xlongequal[]{\text{combine indices}}
\frac{1}{2} \sum_{i \neq k} \frac{\sigma_k}{\sigma_k^2 - \sigma_i^2} \left[ u_i^\top A^{(1)} v_k \right]^2 \\
&\qquad\qquad\qquad\qquad\qquad+
\frac{1}{2} \sum_{i \neq k} \frac{\sigma_k}{\sigma_k^2 - \sigma_i^2} \left[ u_k^\top A^{(1)} v_i \right]^2  \\
&\qquad\qquad\qquad\qquad\qquad +\frac{1}{2}\sum_{i=1}^r \frac{\sigma_i}{\sigma_i^2 - \sigma_k^2} u_i^\top A^{(1)} v_k u_k^\top A^{(1)} v_i \\
&\qquad\qquad\qquad\qquad\qquad +\frac{1}{2}\sum_{i=1}^r \frac{\sigma_i}{\sigma_i^2 - \sigma_k^2} u_k^\top A^{(1)} v_i u_i^\top A^{(1)} v_k  .
\end{align}

\bigskip
\noindent
\textbf{Procedure III -- Map Derivative Operator Layout.} Use following identity from Lemma~\ref{lemma:essential_matrix_identities}~(\nameref{lemma:essential_matrix_identities}):
\begin{enumerate}
    \item $\operatorname{vec}(B V A^\top) = (A \otimes B)\operatorname{vec}(V)$,
\end{enumerate}
consider:
\begin{align}
\lim_{x \to 0} x^2\left[u_i^\top A^{(1)}v_k \right]^2 &=   \lim_{x \to 0} u_i^\top xA^{(1)}v_k u_i^\top xA^{(1)}v_k \\
 &= \left[u_i^\top \dd A v_k\right] \left[u_i^\top \dd A v_k\right] \\
 &=\left[v_k^\top \dd A u_i\right] 
 \left[u_i^\top \dd A v_k\right] \\
 &=\operatorname{vec}\left[v_k^\top \dd A u_i\right]
 \operatorname{vec}\left[u_i^\top \dd A v_k\right] \\
 &=\left(u_i^\top \otimes v_k^\top\right) \operatorname{vec}(\dd A)
 \left(v_k^\top \otimes u_i^\top\right)
 \operatorname{vec}(\dd A) \\
 &=\left[\left(u_i^\top \otimes v_k^\top\right) \operatorname{vec}(\dd A)\right]^\top
 \left[\left(v_k^\top \otimes u_i^\top\right)
 \operatorname{vec}(\dd A)\right] \\
 &=\operatorname{vec}(\dd A^\top)\left(v_k \otimes u_i\right)\left(v_k \otimes u_i\right)^\top\operatorname{vec}(\dd A).
\end{align}

Similarly,
\begin{align}
\lim_{x \to 0} x^2\left[u_k^\top A^{(1)}v_i \right]^2 =    \operatorname{vec}(\dd A^\top)\left(v_i \otimes u_k\right)\left(v_i \otimes u_k\right)^\top\operatorname{vec}(\dd A),
\end{align}
\begin{align}
\lim_{x \to 0} x^2   u_i^\top A^{(1)} v_k u_k^\top A^{(1)} v_i  &= \lim_{x \to 0} u_i^\top xA^{(1)} v_k u_k^\top xA^{(1)} v_i \\
&=\lim_{x \to 0} u_i^\top \dd A v_k u_k^\top \dd A v_i \\
&=\operatorname{vec}(\dd A^\top)\left(v_k \otimes u_i\right)\left(v_i \otimes u_k\right)^\top\operatorname{vec}(\dd A),
\end{align}
and:
\begin{align}
\lim_{x \to 0} x^2   u_k^\top A^{(1)} v_i u_i^\top A^{(1)} v_k  &= \lim_{x \to 0} u_k^\top xA^{(1)} v_i u_i^\top xA^{(1)} v_k \\
&=\lim_{x \to 0} u_k^\top \dd A v_i u_i^\top \dd A v_k \\
&=\operatorname{vec}(\dd A^\top)\left(v_i\otimes u_k\right)\left(v_k \otimes u_i\right)^\top\operatorname{vec}(\dd A)
.
\end{align}

\bigskip
\noindent
\textbf{Producing Lemma Claim.} Hence,
\begin{align}
D^2\sigma_k[\dd A, \dd A]
&=
\operatorname{vec}(\dd A)^\top\frac{\partial}{\partial \operatorname{vec}(A)} \operatorname{vec}\left( \frac{\partial \sigma_k}{\partial A}\right) \operatorname{vec}(\dd A) \\
&=2 \lim_{x \to 0}x^2 \sigma_k^{(2)} \\
&= \sum_{i \neq k} \frac{\sigma_k}{\sigma_k^2 - \sigma_i^2} \lim_{x \to 0} x^2\left[u_i^\top A^{(1)}v_k \right]^2 \\
&\qquad
+ \sum_{i \neq k} \frac{\sigma_k}{\sigma_k^2 - \sigma_i^2} \lim_{x \to 0} x^2\left[u_k^\top A^{(1)}v_i \right]^2 \\
&\qquad
+\sum_{i=1}^r \frac{\sigma_i}{\sigma_i^2 - \sigma_k^2} \lim_{x \to 0} x^2   u_i^\top A^{(1)} v_k u_k^\top A^{(1)} v_i \\
&\qquad
+\sum_{i=1}^r \frac{\sigma_i}{\sigma_i^2 - \sigma_k^2} \lim_{x \to 0} x^2   u_k^\top A^{(1)} v_i u_i^\top A^{(1)} v_k 
.
\end{align}

Re-labeling indices yields the claim:\\
\pgfmathsetmacro{\linewidthfactor}{0.9}
\pgfmathsetmacro{\inverselinewidthfactor}{1/\linewidthfactor}
\scalebox{\linewidthfactor}{% scale
\begin{minipage}{\inverselinewidthfactor\linewidth} % 1/scale
\begin{align}
D^2\sigma_k[\dd A, \dd A] &=
\operatorname{vec}(\dd A)^\top
\left[
\underbrace{\sum_{i \neq k, i \leq r} \frac{\sigma_k}{\sigma_k^2 - \sigma_i^2}\left(v_k \otimes u_i\right)\left(v_k \otimes u_i\right)^\top}_{\text{left}} \right.\\
&\left.\qquad\qquad\qquad
+\underbrace{\sum_{j \neq k, j \leq n} \frac{\sigma_k}{\sigma_k^2 - \sigma_j^2} \left(v_j \otimes u_k\right)\left(v_j \otimes u_k\right)^\top}_{\text{right}}  \right.\\
    &\left.\qquad\qquad \qquad+\underbrace{\sum_{l \neq k, l \leq r} \frac{\sigma_l}{\sigma_k^2 - \sigma_l^2}\left[ \left(v_k \otimes u_l\right)\left(v_l \otimes u_k\right)^\top + \left(v_l \otimes u_k\right)\left(v_k \otimes u_l\right)^\top \right]}_{\text{left-right interaction}} \right]
    \operatorname{vec}(\dd A)  
    .
\end{align}
\end{minipage}
}\\

%\end{proofpart}

{\hfill$\qed$\par}
\end{mynewproof}

\begin{figure}%[t!]
  \centering

  \begin{minipage}[t]{1\textwidth}
     \centering 
     \subfloat[{Matrix entries i.i.d. $\sim \mathcal{N}(0, 1)$.}]{%
    \centering
    \includegraphics[width=0.8\linewidth]{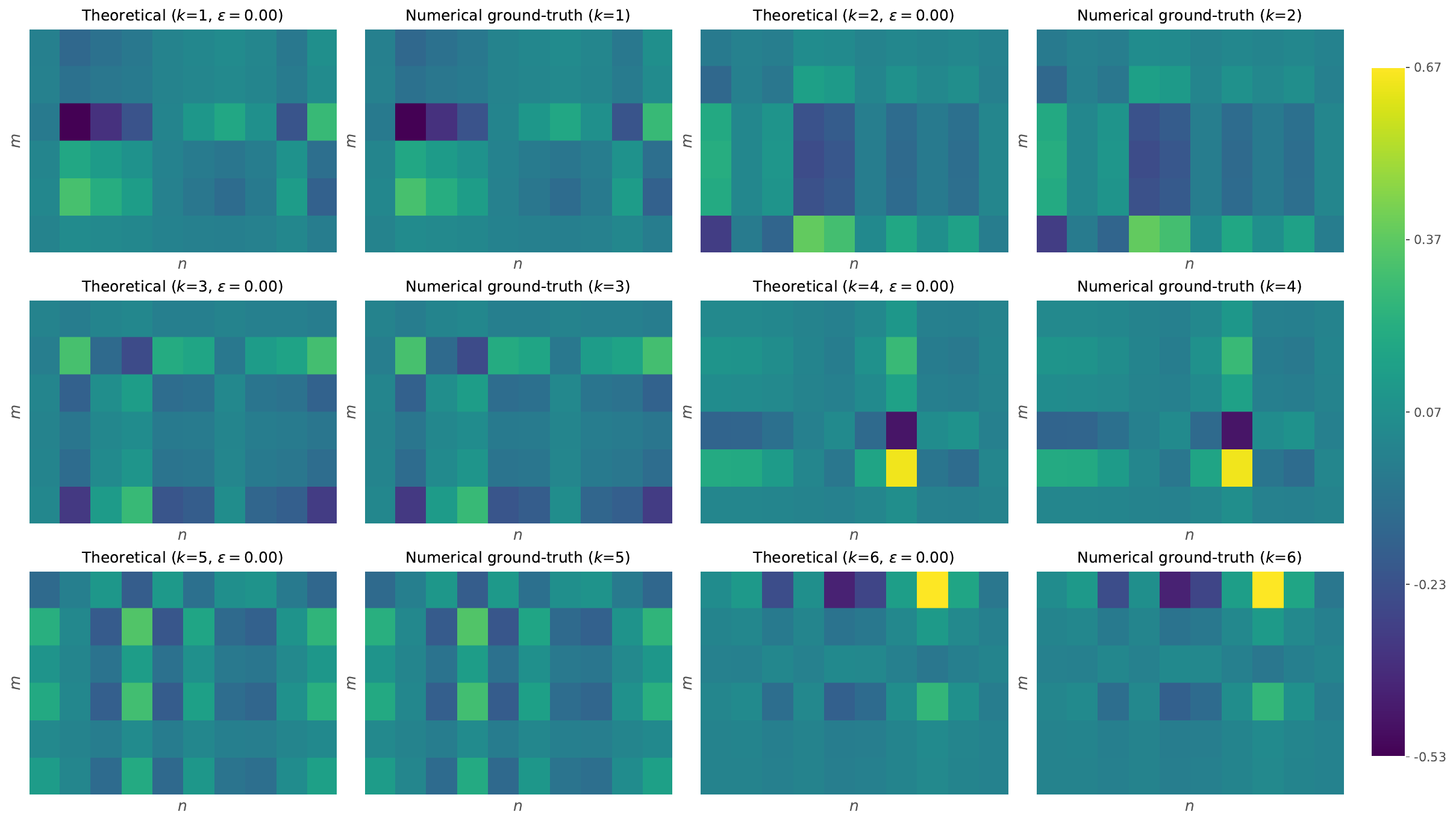}%
    %\caption{The matrix is sampled from $\mathcal{N}(0, 1)$.}
    \label{fig:hessian:normal}
  }
  \end{minipage}
  
  \vspace{1em}
  
  \begin{minipage}[t]{1\textwidth}
     \centering 
   \subfloat[{Matrix entries i.i.d. $\sim U[0,1]$.}]{%
    \centering
    \includegraphics[width=0.8\linewidth]{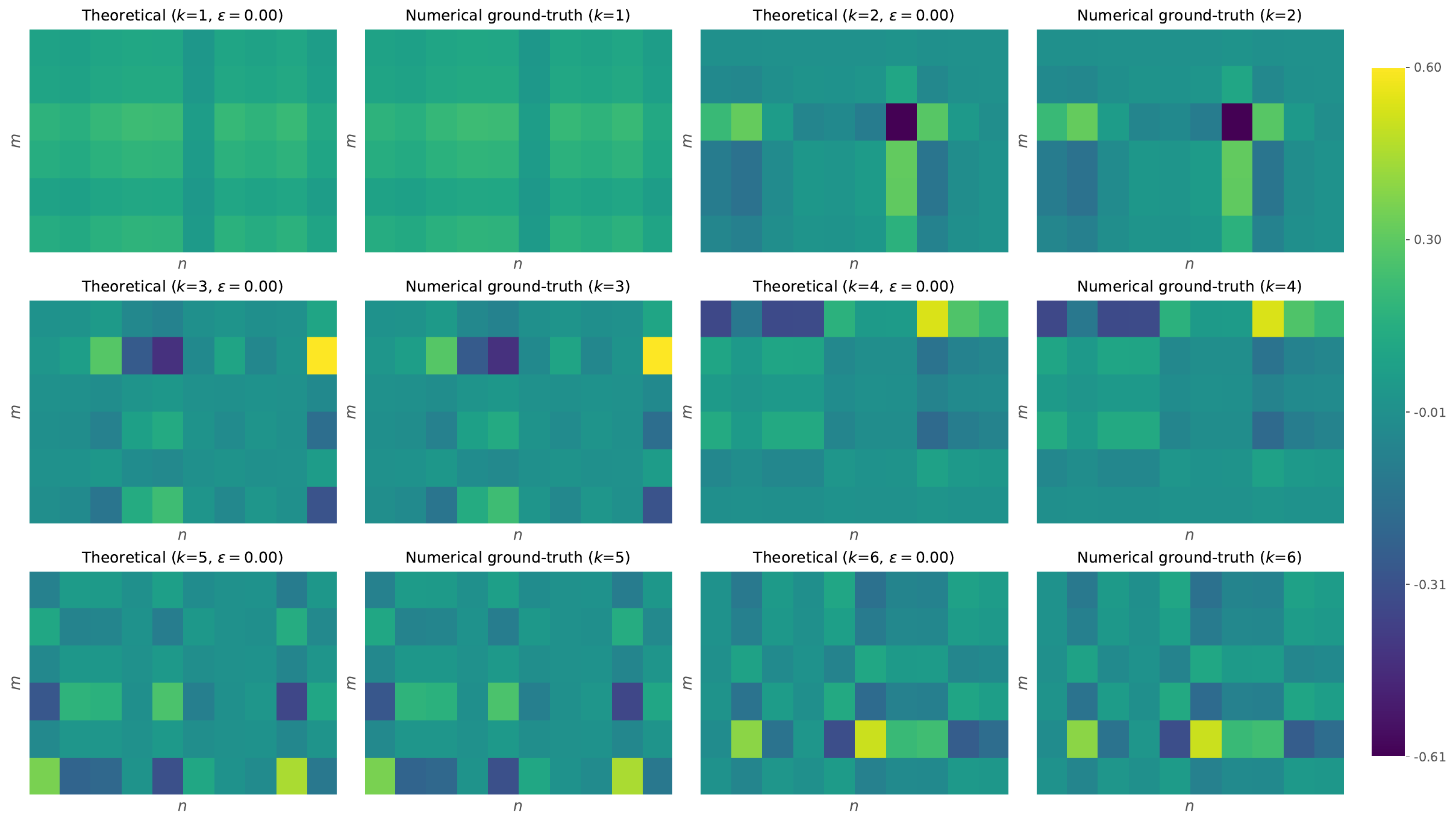}%
    %\caption{The matrix is sampled from $U[0, 1]$.}
    \label{fig:jacobian:uniform}
  }
  \end{minipage}
  
  \caption[Numerical Experiments for Singular-Value Jacobian]{\textbf{Numerical Experiments for Singular-Value Jacobian.} This experiment compares the singular-value Jacobian derived from our framework with that obtained via PyTorch's auto--differentiation. The error $\epsilon$ is measured as the $\ell_2$-norm between the theoretical and ground-truth results. The error is measured to be zero in these experiments, indicating no difference between the theoretical and ground-truth results.}
  
  \label{fig:jacobian:all}
  
\end{figure}

\section{Numerical Experiments}
\label{sec:numerical_experiments}

\begin{figure}%[t!]
  \centering

    \begin{minipage}[t]{1\textwidth}
     \centering 
     \subfloat[{Matrix entries i.i.d. $\sim \mathcal{N}(0,1)$.}]{%
    \centering
    \includegraphics[width=0.8\linewidth]{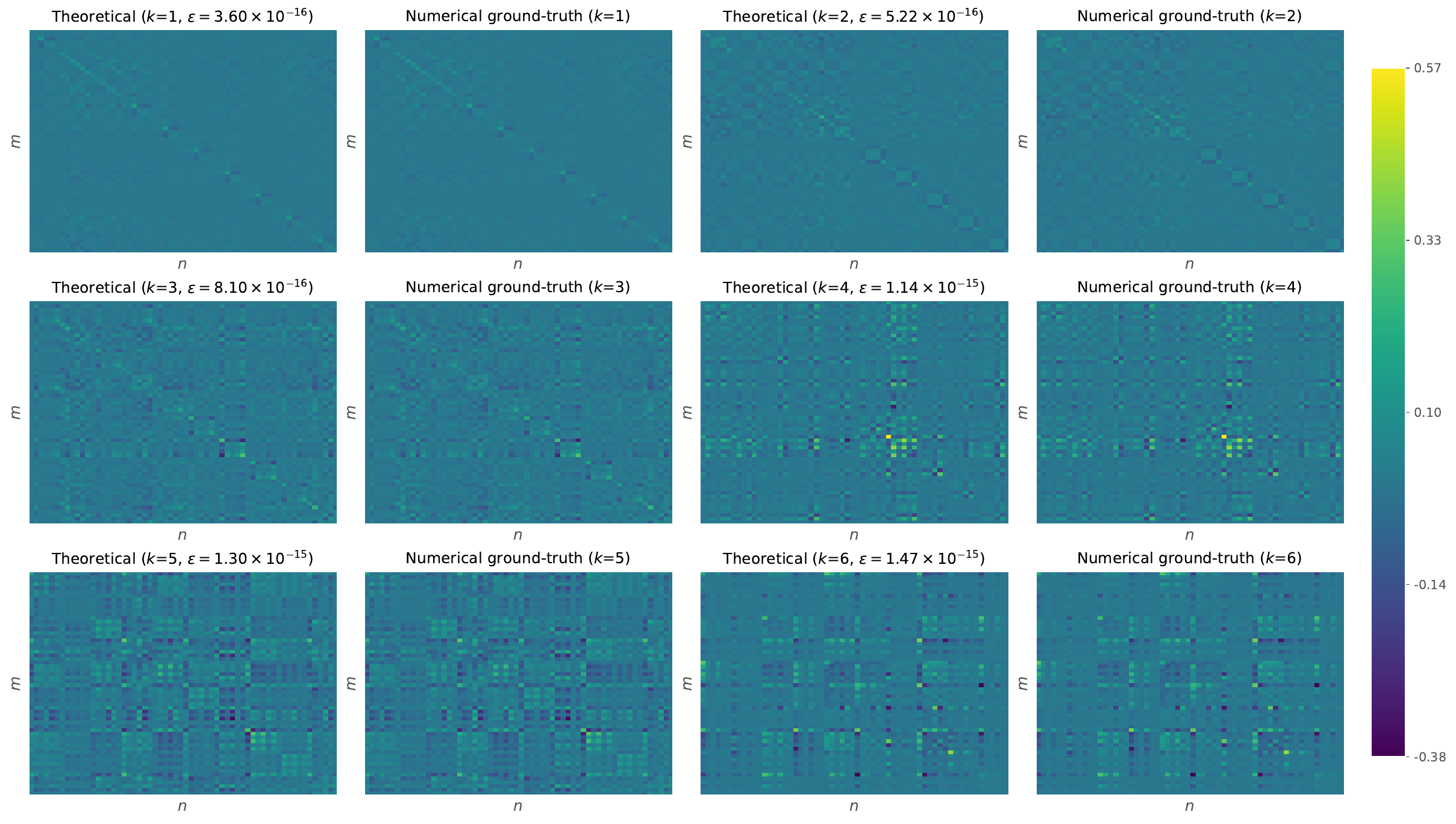}
    %\caption{The matrix is sampled from $\mathcal{N}(0, 1)$.}
    \label{fig:hessian:normal}
  }
  \end{minipage}
  
  %\vspace{1em}
  
\begin{minipage}[t]{1\textwidth}
     \centering 
     \subfloat[{Matrix entries i.i.d. $\sim U[0, 1]$.}]{%
    \centering
    \includegraphics[width=0.8\linewidth]{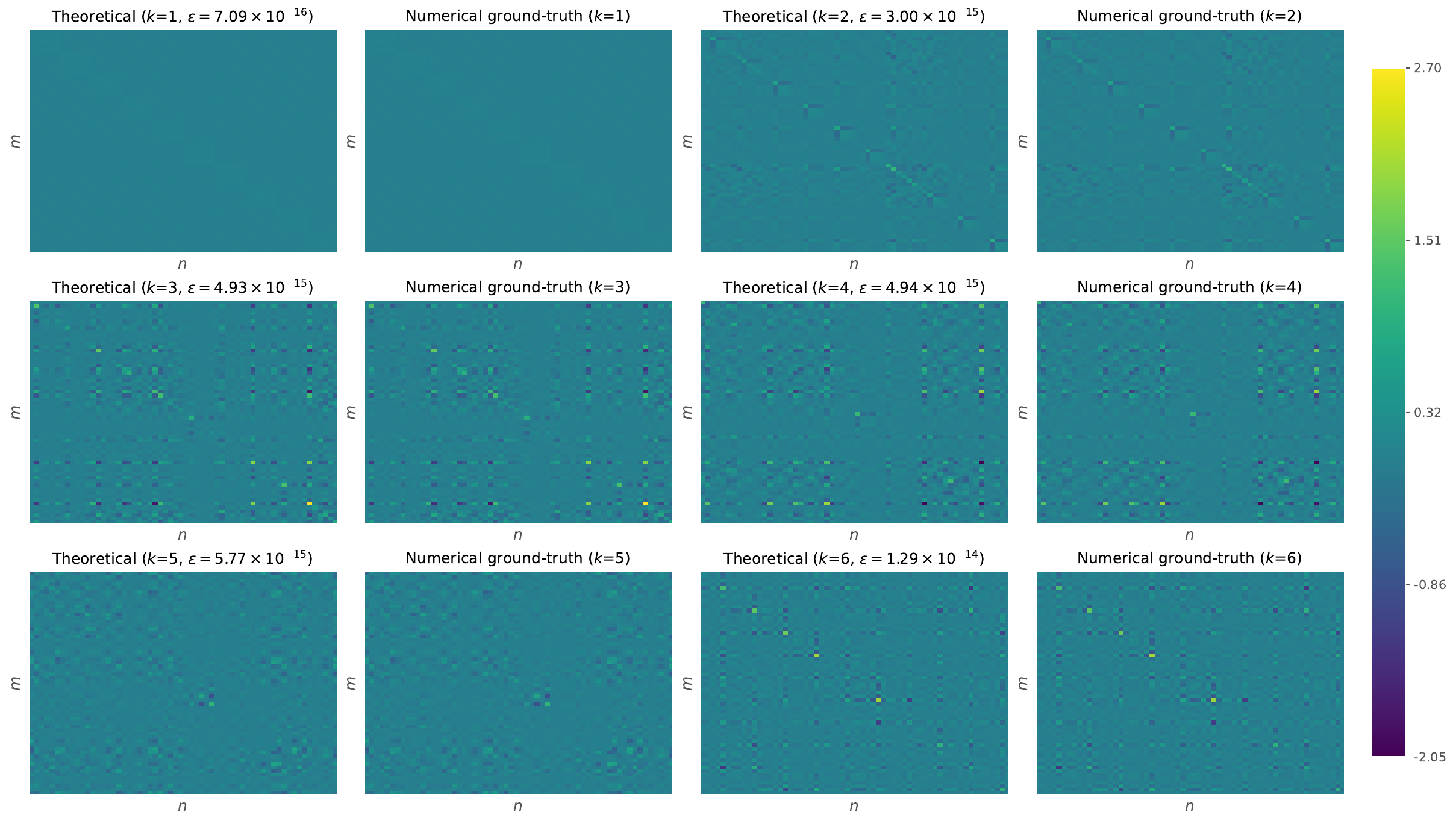}
    %\caption{The matrix is sampled from $U[0, 1]$.}
    \label{fig:hessian:uniform}
  }
  \end{minipage}

  \caption[Numerical Experiments for Singular-Value Hessian]{\textbf{Numerical Experiments for Singular-Value Hessian.} This experiment compares the singular-value Hessian derived from our framework with that obtained via PyTorch's auto--differentiation. The error $\epsilon$ is measured as the $\ell_2$-norm between the theoretical and ground-truth results. The maximum error is measured to be less than $1.3\times 10^{-14}$ in these experiments, indicating the difference between the theoretical and ground-truth results is negligible.}
  
  \label{fig:hessian:all}
\end{figure}

\begin{figure}%[t!]
  \centering

    \begin{minipage}[t]{0.48\textwidth}
     \centering 
     \subfloat[{Matrix entries i.i.d. $\sim \mathcal{N}(0,1)$.}]{%
    \centering
    \includegraphics[width=0.9\linewidth]{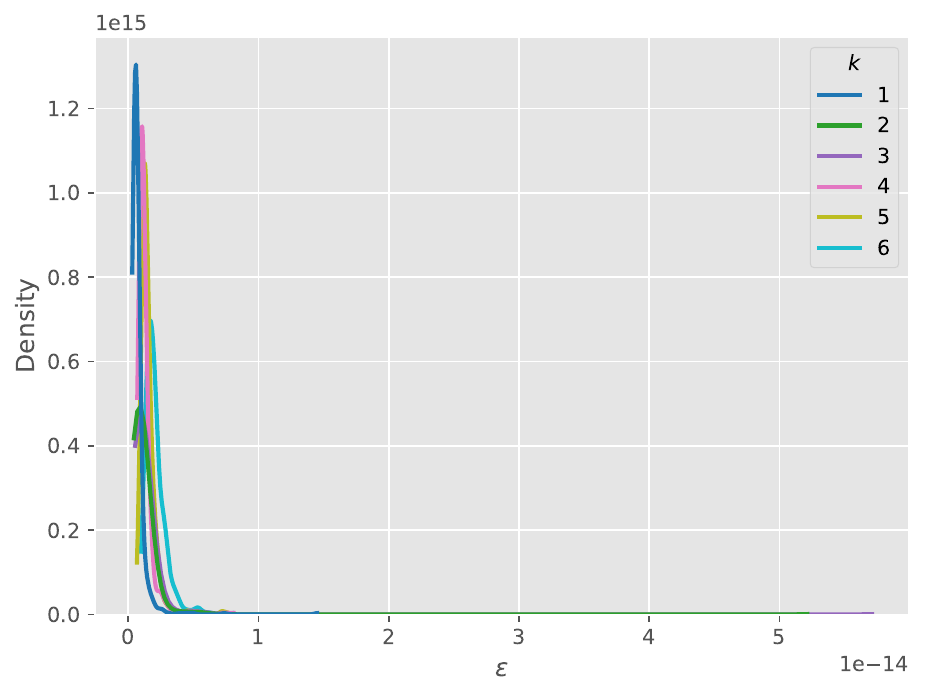}
    \label{fig:hessian:errors:normal}
  }
  \end{minipage}
\begin{minipage}[t]{0.48\textwidth}
     \centering 
     \subfloat[{Matrix entries i.i.d. $\sim U[0, 1]$.}]{%
    \centering
    \includegraphics[width=0.9\linewidth]{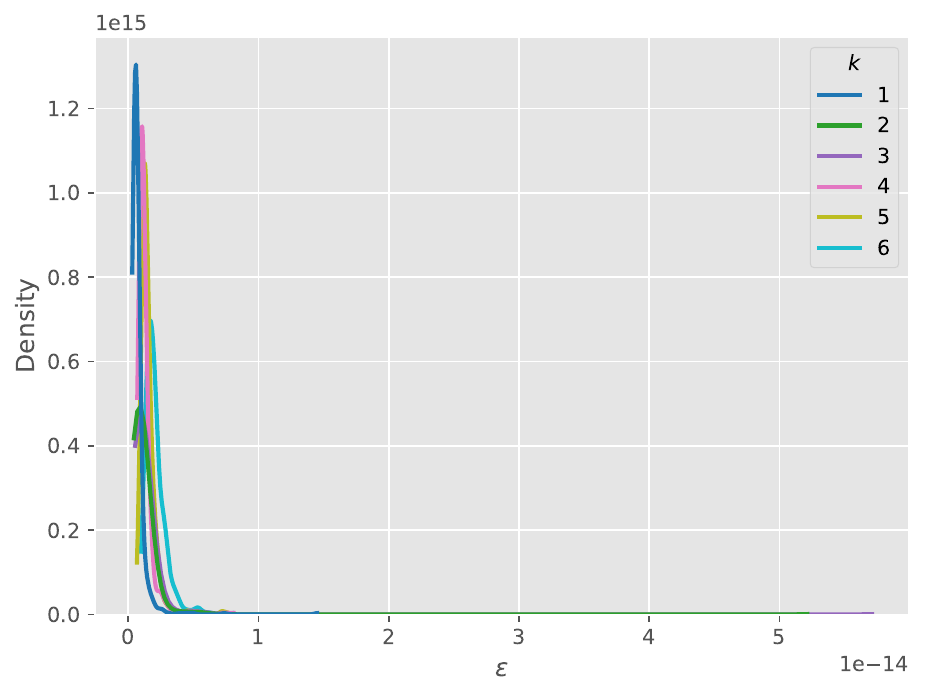}
    \label{fig:hessian:errors:uniform}
  }
  \end{minipage}

  \caption[Errors for Singular-Value Hessian]{\textbf{Errors for Singular-Value Hessian.} Random matrix entries are sampled i.i.d.\ from $\mathcal{N}(0,1)$ and $U[0,1]$, respectively. For each singular-value index $k=1,2,\dots,r$, the error $\epsilon$ is computed over $500$ trials and visualized using an unnormalized histogram density. All reported errors are below $6\times 10^{-14}$ in these experiments.}
  
  \label{fig:hessian:errors}
\end{figure}

We conduct numerical experiments to validate the correctness of the derived special cases $n=1$ and $n=2$. Matrix entries are sampled from $\mathcal{N}(0,1)$ and $U(0,1)$, respectively. Ground truth is obtained numerically via PyTorch's auto--differentiation framework \citep{paszke2019pytorch}. The error $\epsilon$ is computed by the $\ell_2$-norm
\begin{align}
    \epsilon = \| R_{\text{ours}} - R_{\text{gt}} \|_2,
\end{align}
where $R_{\text{ours}}$ denotes the result from our theoretical computation and $R_{\text{gt}}$ the ground truth from auto--differentiation. Singular values are indexed by $k=1,2,\dots,r$ in the reported results. To facilitate the visualization and computation, we choose the dimensions $6 \times 10$ in all experiments. 

\bigskip
\noindent
\textbf{Reproducibility.} The random seed is fixed to $1$ for reproducibility. All experimental code is available at \url{https://github.com/roisincrtai/highorder_spectral_variation_analysis}.

\bigskip
\noindent
\textbf{Results of Singular-Value Jacobian.} Figure~\ref{fig:jacobian:all} reports the results for the singular-value Jacobian. Matrix entries are sampled i.i.d. from $\mathcal{N}(0,1)$ and $U(0,1)$, respectively. The derivative matrices are visualized using the \emph{viridis} color map. For each singular-value index $k$, results are shown in pairs: the left panel gives the theoretical computation from Lemma~\ref{lemma:jacobian_of_singular_value}, while the right panel shows the numerical ground truth obtained from PyTorch's auto--differentiation framework. The reported errors are zero across all experiments.

\bigskip
\noindent
\textbf{Results of Singular-Value Hessian.} 
Figure~\ref{fig:hessian:all} reports the results for the singular-value Hessian. Matrix entries are sampled i.i.d. from $\mathcal{N}(0,1)$ and $U(0,1)$, respectively. The derivative matrices are visualized using the \emph{viridis} color map. For each singular-value index $k$, results are shown in pairs: the left panel gives the theoretical computation from Lemma~\ref{lemma:hessian_of_singular_value}, while the right panel shows the numerical ground truth obtained from PyTorch's auto--differentiation framework. The observed errors between theoretical results and numerical ground-truth are on the order of $10^{-14}$, confirming that they are numerically negligible.

\section{Conclusion}

By viewing matrices as compact linear operators and extending Kato's perturbation theory for self-adjoint operators, we present a unified operator-theoretic framework for obtaining closed-form, arbitrary-order derivatives of singular values in real rectangular matrices. In contrast to the ad hoc methods of classical matrix analysis, our approach is systematic and procedural, allowing the derivation of singular-value derivatives of any order. The key step is the Jordan–Wielandt embedding, which maps a real rectangular matrix, usually non-self-adjoint, to a self-adjoint operator, thereby encapsulating its complete spectral information. Based on Kato's framework, we establish a general framework for deriving higher-order singular-value derivatives. Specializing to first order ($n=1$) recovers the classical singular-value Jacobian, while specializing to second order ($n=2$) yields a Kronecker-product representation of the singular-value Hessian that, to the best of our knowledge, has not previously appeared in the literature. Beyond these cases, the framework extends to arbitrary order. Higher-order singular-value derivatives are indispensable for analyzing induced spectral dynamics in statistical physics and deep learning.

\section*{Acknowledgments}
This publication has emanated from research conducted with the financial support of \textbf{Taighde \'Eireann} - Research Ireland under Grant number 18/CRT/6223. %We also thank Prof. Dr. 

%%%%%%%%%%%%%%%%%%%%%%%%%%%%%%%%%%%%%%%%%%%%%%%%%%%%%%%%%%%%

% \appendix
% \input{appendix}

% \section{Results of Numerical Experiments}\label{}


\bibliographystyle{cas-model2-names}

\bibliography{references}

\begin{thebibliography}{37}
\expandafter\ifx\csname natexlab\endcsname\relax\def\natexlab#1{#1}\fi
\providecommand{\url}[1]{\texttt{#1}}
\providecommand{\href}[2]{#2}
\providecommand{\path}[1]{#1}
\providecommand{\DOIprefix}{doi:}
\providecommand{\ArXivprefix}{arXiv:}
\providecommand{\URLprefix}{URL: }
\providecommand{\Pubmedprefix}{pmid:}
\providecommand{\doi}[1]{\href{http://dx.doi.org/#1}{\path{#1}}}
\providecommand{\Pubmed}[1]{\href{pmid:#1}{\path{#1}}}
\providecommand{\bibinfo}[2]{#2}
\ifx\xfnm\relax \def\xfnm[#1]{\unskip,\space#1}\fi
%Type = Book
\bibitem[{Applebaum(2009)}]{applebaum2009levy}
\bibinfo{author}{Applebaum, D.}, \bibinfo{year}{2009}.
\newblock \bibinfo{title}{L{\'e}vy processes and stochastic calculus}.
\newblock \bibinfo{publisher}{Cambridge university press}.
%Type = Book
\bibitem[{Bai et~al.(2010)Bai, Silverstein et~al.}]{bai2010spectral}
\bibinfo{author}{Bai, Z.}, \bibinfo{author}{Silverstein, J.W.}, et~al.,
  \bibinfo{year}{2010}.
\newblock \bibinfo{title}{Spectral analysis of large dimensional random
  matrices}.
\newblock \bibinfo{publisher}{Springer}.
%Type = Book
\bibitem[{Bhatia(2013)}]{bhatia2013matrix}
\bibinfo{author}{Bhatia, R.}, \bibinfo{year}{2013}.
\newblock \bibinfo{title}{Matrix analysis}. volume \bibinfo{volume}{169}.
\newblock \bibinfo{publisher}{Springer Science \& Business Media}.
%Type = Book
\bibitem[{Clarke(1990)}]{clarke1990optimization}
\bibinfo{author}{Clarke, F.H.}, \bibinfo{year}{1990}.
\newblock \bibinfo{title}{Optimization and nonsmooth analysis}.
\newblock \bibinfo{publisher}{SIAM}.
%Type = Book
\bibitem[{Dunford and Schwartz(1988)}]{dunford1988linear}
\bibinfo{author}{Dunford, N.}, \bibinfo{author}{Schwartz, J.T.},
  \bibinfo{year}{1988}.
\newblock \bibinfo{title}{Linear operators, part 1: general theory}.
\newblock \bibinfo{publisher}{John Wiley \& Sons}.
%Type = Article
\bibitem[{Edelman and Rao(2005)}]{edelman2005random}
\bibinfo{author}{Edelman, A.}, \bibinfo{author}{Rao, N.R.},
  \bibinfo{year}{2005}.
\newblock \bibinfo{title}{Random matrix theory}.
\newblock \bibinfo{journal}{Acta numerica} \bibinfo{volume}{14},
  \bibinfo{pages}{233--297}.
%Type = Article
\bibitem[{Fano(1957)}]{fano1957description}
\bibinfo{author}{Fano, U.}, \bibinfo{year}{1957}.
\newblock \bibinfo{title}{Description of states in quantum mechanics by density
  matrix and operator techniques}.
\newblock \bibinfo{journal}{Reviews of modern physics} \bibinfo{volume}{29},
  \bibinfo{pages}{74}.
%Type = Book
\bibitem[{Franklin(2000)}]{franklin2000matrix}
\bibinfo{author}{Franklin, J.N.}, \bibinfo{year}{2000}.
\newblock \bibinfo{title}{Matrix theory}.
\newblock \bibinfo{publisher}{Courier Corporation}.
%Type = Book
\bibitem[{Horn and Johnson(2012)}]{Horn2012}
\bibinfo{author}{Horn, R.A.}, \bibinfo{author}{Johnson, C.R.},
  \bibinfo{year}{2012}.
\newblock \bibinfo{title}{Matrix analysis}.
\newblock \bibinfo{publisher}{Cambridge university press}.
%Type = Article
\bibitem[{It\^o(1951)}]{ito1951stochastic}
\bibinfo{author}{It\^o, K.}, \bibinfo{year}{1951}.
\newblock \bibinfo{title}{On stochastic differential equations}.
\newblock \bibinfo{journal}{Memoirs of the American Mathematical Society} ,
  \bibinfo{pages}{1--51}.
%Type = Book
\bibitem[{Kato(1995)}]{kato1995perturbation}
\bibinfo{author}{Kato, T.}, \bibinfo{year}{1995}.
\newblock \bibinfo{title}{Perturbation theory for linear operators}.
\newblock \bibinfo{publisher}{Springer}.
%Type = Article
\bibitem[{Kolda and Bader(2009)}]{kolda2009tensor}
\bibinfo{author}{Kolda, T.G.}, \bibinfo{author}{Bader, B.W.},
  \bibinfo{year}{2009}.
\newblock \bibinfo{title}{Tensor decompositions and applications}.
\newblock \bibinfo{journal}{SIAM review} \bibinfo{volume}{51},
  \bibinfo{pages}{455--500}.
%Type = Article
\bibitem[{Lewis and Sendov(2005)}]{lewis2005nonsmooth}
\bibinfo{author}{Lewis, A.S.}, \bibinfo{author}{Sendov, H.S.},
  \bibinfo{year}{2005}.
\newblock \bibinfo{title}{Nonsmooth analysis of singular values. part i:
  Theory}.
\newblock \bibinfo{journal}{Set-Valued Analysis} \bibinfo{volume}{13},
  \bibinfo{pages}{213--241}.
%Type = Article
\bibitem[{Li and Li(2005)}]{li2005note}
\bibinfo{author}{Li, C.K.}, \bibinfo{author}{Li, R.C.}, \bibinfo{year}{2005}.
\newblock \bibinfo{title}{A note on eigenvalues of perturbed hermitian
  matrices}.
\newblock \bibinfo{journal}{Linear algebra and its applications}
  \bibinfo{volume}{395}, \bibinfo{pages}{183--190}.
%Type = Article
\bibitem[{Liu et~al.(2024)Liu, Trenkler, Kollo, von Rosen and
  Baksalary}]{liu2024professor}
\bibinfo{author}{Liu, S.}, \bibinfo{author}{Trenkler, G.},
  \bibinfo{author}{Kollo, T.}, \bibinfo{author}{von Rosen, D.},
  \bibinfo{author}{Baksalary, O.M.}, \bibinfo{year}{2024}.
\newblock \bibinfo{title}{Professor heinz neudecker and matrix differential
  calculus}.
\newblock \bibinfo{journal}{Statistical Papers} \bibinfo{volume}{65},
  \bibinfo{pages}{2605--2639}.
%Type = Article
\bibitem[{Luo et~al.(2025)Luo, McDermott, Gagn{\'e}, Sun and
  O'Riordan}]{luo2025lipschitz}
\bibinfo{author}{Luo, R.}, \bibinfo{author}{McDermott, J.},
  \bibinfo{author}{Gagn{\'e}, C.}, \bibinfo{author}{Sun, Q.},
  \bibinfo{author}{O'Riordan, C.}, \bibinfo{year}{2025}.
\newblock \bibinfo{title}{Optimization-induced dynamics of lipschitz continuity
  in neural networks}.
\newblock \bibinfo{journal}{arXiv preprint arXiv:2506.18588} .
%Type = Book
\bibitem[{Magnus and Neudecker(2019)}]{magnusmatrix}
\bibinfo{author}{Magnus, J.R.}, \bibinfo{author}{Neudecker, H.},
  \bibinfo{year}{2019}.
\newblock \bibinfo{title}{Matrix differential calculus with applications in
  statistics and econometrics}.
\newblock \bibinfo{publisher}{John Wiley \& Sons}.
%Type = Article
\bibitem[{Mar{\v{c}}enko and Pastur(1967)}]{MarcenkoPastur1967}
\bibinfo{author}{Mar{\v{c}}enko, V.A.}, \bibinfo{author}{Pastur, L.A.},
  \bibinfo{year}{1967}.
\newblock \bibinfo{title}{Distribution of eigenvalues for some sets of random
  matrices}.
\newblock \bibinfo{journal}{Mathematics of the USSR-Sbornik}
  \bibinfo{volume}{1}, \bibinfo{pages}{457}.
%Type = Book
\bibitem[{Mehta(2004)}]{Mehta2004}
\bibinfo{author}{Mehta, M.L.}, \bibinfo{year}{2004}.
\newblock \bibinfo{title}{Random matrices}. volume \bibinfo{volume}{142}.
\newblock \bibinfo{publisher}{Elsevier}.
%Type = Book
\bibitem[{Nielsen and Chuang(2010)}]{nielsen2010quantum}
\bibinfo{author}{Nielsen, M.A.}, \bibinfo{author}{Chuang, I.L.},
  \bibinfo{year}{2010}.
\newblock \bibinfo{title}{Quantum computation and quantum information}.
\newblock \bibinfo{publisher}{Cambridge university press}.
%Type = Book
\bibitem[{Oksendal(2013)}]{oksendal2003stochastic}
\bibinfo{author}{Oksendal, B.}, \bibinfo{year}{2013}.
\newblock \bibinfo{title}{Stochastic differential equations: an introduction
  with applications}.
\newblock \bibinfo{publisher}{Springer Science \& Business Media}.
%Type = Article
\bibitem[{Paszke et~al.(2019)Paszke, Gross, Massa, Lerer, Bradbury, Chanan,
  Killeen, Lin, Gimelshein, Antiga et~al.}]{paszke2019pytorch}
\bibinfo{author}{Paszke, A.}, \bibinfo{author}{Gross, S.},
  \bibinfo{author}{Massa, F.}, \bibinfo{author}{Lerer, A.},
  \bibinfo{author}{Bradbury, J.}, \bibinfo{author}{Chanan, G.},
  \bibinfo{author}{Killeen, T.}, \bibinfo{author}{Lin, Z.},
  \bibinfo{author}{Gimelshein, N.}, \bibinfo{author}{Antiga, L.}, et~al.,
  \bibinfo{year}{2019}.
\newblock \bibinfo{title}{Pytorch: An imperative style, high-performance deep
  learning library}.
\newblock \bibinfo{journal}{Advances in neural information processing systems}
  \bibinfo{volume}{32}.
%Type = Book
\bibitem[{Rayleigh(1896)}]{rayleigh1896theory}
\bibinfo{author}{Rayleigh, J.W.S.B.}, \bibinfo{year}{1896}.
\newblock \bibinfo{title}{The theory of sound}. volume~\bibinfo{volume}{2}.
\newblock \bibinfo{publisher}{Macmillan}.
%Type = Book
\bibitem[{Rellich(1969)}]{rellich1969perturbation}
\bibinfo{author}{Rellich, F.}, \bibinfo{year}{1969}.
\newblock \bibinfo{title}{Perturbation theory of eigenvalue problems}.
\newblock \bibinfo{publisher}{CRC Press}.
%Type = Book
\bibitem[{Rudin(1991)}]{Rudin1991}
\bibinfo{author}{Rudin, W.}, \bibinfo{year}{1991}.
\newblock \bibinfo{title}{Functional Analysis}.
\newblock \bibinfo{edition}{2nd} ed.
%Type = Book
\bibitem[{Sakurai and Napolitano(2020)}]{sakurai2020modern}
\bibinfo{author}{Sakurai, J.J.}, \bibinfo{author}{Napolitano, J.},
  \bibinfo{year}{2020}.
\newblock \bibinfo{title}{Modern quantum mechanics}.
\newblock \bibinfo{publisher}{Cambridge University Press}.
%Type = Article
\bibitem[{Schr{\"o}dinger(1926)}]{schrodinger1926quantisierung}
\bibinfo{author}{Schr{\"o}dinger, E.}, \bibinfo{year}{1926}.
\newblock \bibinfo{title}{Quantisierung als eigenwertproblem}.
\newblock \bibinfo{journal}{Annalen der physik} \bibinfo{volume}{385},
  \bibinfo{pages}{437--490}.
%Type = Inproceedings
\bibitem[{Shalit(2021)}]{shalit2021dilation}
\bibinfo{author}{Shalit, O.M.}, \bibinfo{year}{2021}.
\newblock \bibinfo{title}{Dilation theory: a guided tour}, in:
  \bibinfo{booktitle}{Operator theory, functional analysis and applications},
  \bibinfo{organization}{Springer}. pp. \bibinfo{pages}{551--623}.
%Type = Book
\bibitem[{Spivak(2018)}]{spivak2018calculus}
\bibinfo{author}{Spivak, M.}, \bibinfo{year}{2018}.
\newblock \bibinfo{title}{Calculus on manifolds: a modern approach to classical
  theorems of advanced calculus}.
\newblock \bibinfo{publisher}{CRC press}.
%Type = Article
\bibitem[{Stewart and Sun(1990)}]{StewartSun1990}
\bibinfo{author}{Stewart, G.W.}, \bibinfo{author}{Sun, J.g.},
  \bibinfo{year}{1990}.
\newblock \bibinfo{title}{Matrix perturbation theory}.
\newblock \bibinfo{journal}{Academic Press} .
%Type = Book
\bibitem[{Strang(2012)}]{strang2012linear}
\bibinfo{author}{Strang, G.}, \bibinfo{year}{2012}.
\newblock \bibinfo{title}{Linear algebra and its applications}.
%Type = Book
\bibitem[{Tao(2012)}]{tao2012topics}
\bibinfo{author}{Tao, T.}, \bibinfo{year}{2012}.
\newblock \bibinfo{title}{Topics in random matrix theory}. volume
  \bibinfo{volume}{132}.
\newblock \bibinfo{publisher}{American Mathematical Soc.}
%Type = Article
\bibitem[{Tracy and Widom(1994)}]{TracyWidom1994}
\bibinfo{author}{Tracy, C.A.}, \bibinfo{author}{Widom, H.},
  \bibinfo{year}{1994}.
\newblock \bibinfo{title}{Level-spacing distributions and the airy kernel}.
\newblock \bibinfo{journal}{Communications in Mathematical Physics}
  \bibinfo{volume}{159}, \bibinfo{pages}{151--174}.
%Type = Article
\bibitem[{Wedin(1972)}]{Wedin1972}
\bibinfo{author}{Wedin, P.{\AA}.}, \bibinfo{year}{1972}.
\newblock \bibinfo{title}{Perturbation bounds in connection with singular value
  decomposition}.
\newblock \bibinfo{journal}{BIT Numerical Mathematics} \bibinfo{volume}{12},
  \bibinfo{pages}{99--111}.
%Type = Article
\bibitem[{Weyl(1912)}]{weyl1912asymptotische}
\bibinfo{author}{Weyl, H.}, \bibinfo{year}{1912}.
\newblock \bibinfo{title}{Das asymptotische verteilungsgesetz der eigenwerte
  linearer partieller differentialgleichungen (mit einer anwendung auf die
  theorie der hohlraumstrahlung)}.
\newblock \bibinfo{journal}{Mathematische Annalen} \bibinfo{volume}{71},
  \bibinfo{pages}{441--479}.
%Type = Article
\bibitem[{Wielandt et~al.(1955)}]{wielandt1955eigenvalues}
\bibinfo{author}{Wielandt, H.}, et~al., \bibinfo{year}{1955}.
\newblock \bibinfo{title}{On eigenvalues of sums of normal matrices}.
\newblock \bibinfo{journal}{Pacific J. Math} \bibinfo{volume}{5},
  \bibinfo{pages}{633--638}.
%Type = Book
\bibitem[{Yosida(2012)}]{yosida2012functional}
\bibinfo{author}{Yosida, K.}, \bibinfo{year}{2012}.
\newblock \bibinfo{title}{Functional analysis}. volume \bibinfo{volume}{123}.
\newblock \bibinfo{publisher}{Springer Science \& Business Media}.

\end{thebibliography}

%  Here goes the biography details.


\end{document}